\documentclass[Afour,sageh,times]{sagej}

\usepackage{moreverb,url}

\usepackage[colorlinks,bookmarksopen,bookmarksnumbered,citecolor=red,urlcolor=red]{hyperref}

\newcommand\numberthis{\addtocounter{equation}{1}\tag{\theequation}}
\DeclareMathOperator*{\argmin}{argmin}
\DeclareMathOperator*{\argmax}{argmax}

\usepackage{algorithmic}
\usepackage{algorithm}
\usepackage{booktabs}
\usepackage{multirow}
\usepackage{mathtools}
\mathtoolsset{showonlyrefs=true}
\newtheorem{example}{Example}
\usepackage{xcolor}

\newcommand{\ignore}[1]{}

\newcommand{\norm}[1]{\left\Vert#1\right\Vert} 
\newcommand{\mc}[1]{\mathcal{#1}}  

\newcommand{\frob}{{\ensuremath{\mathsf{F}}}}

\newcommand{\bma}[1]{\left[\begin{array}{ #1}}
\newcommand{\ema}{\end{array}\right]}

\DeclareMathAlphabet{\mbf}{OT1}{ptm}{b}{n}
\newcommand{\mbs}[1]{{\boldsymbol{#1}}}
\newcommand{\mbc}[1]{ \boldsymbol{\mathcal{#1}} } 

\newcommand{\mbfbar}[1]{{\bar{\mbf{#1}}}}
\newcommand{\mbfhat}[1]{{\hat{\mbf{#1}}}}
\newcommand{\mbfcheck}[1]{{\check{\mbf{#1}}}}

\newcommand{\mbftilde}[1]{{\tilde{\mbf{#1}}}}

\newcommand{\Exp}{\mathrm{Exp}}
\newcommand{\Log}{\mathrm{Log}}
\newcommand{\mchat}[1]{\hat{\mc{#1}}}
\newcommand{\mccheck}[1]{\check{\mc{#1}}}
\newcommand{\mctilde}[1]{\tilde{\mc{#1}}}

\def\fdotb{{\raisebox{-0.6ex}{ \kern0.2ex\raisebox{0.8ex}{\tiny $\hspace*{-1ex}\circ$}}}}
\def\fddotb{{\raisebox{-0.6ex}{ \kern0.2ex\raisebox{0.8ex}{\tiny $\hspace*{-1ex}\circ\circ$}}}}

\newtheorem{lemma}{Lemma} 

\newcommand{\p}{\partial}

\newcommand{\trans}{{\ensuremath{\mathsf{T}}}} 
\newcommand{\utimes}{ {\raisebox{-0.6ex}{ \kern-1.0ex\raisebox{0.6ex}{ \small $\mathsf{v}$}}} } %
\newcommand{\beq}{\begin{equation}}
\newcommand{\eeq}{\end{equation}}
\newcommand{\bdis}{\begin{displaymath}}
\newcommand{\edis}{\end{displaymath}}
\newcommand{\beqarray}{\begin{eqnarray}}
\newcommand{\eeqarray}{\end{eqnarray}}
\newcommand{\beqarraynn}{\begin{eqnarray*}}
\newcommand{\eeqarraynn}{\end{eqnarray*}}
\newcommand{\balign}{\begin{align}}
\newcommand{\ealign}{\end{align}}
\newcommand{\balignnn}{\begin{align*}}
\newcommand{\ealignnn}{\end{align}}

\makeatletter
\renewcommand{\p@enumii}{\theenumi.}
\makeatother

\newcommand\BibTeX{{\rmfamily B\kern-.05em \textsc{i\kern-.025em b}\kern-.08em
T\kern-.1667em\lower.7ex\hbox{E}\kern-.125emX}}

\newcommand{\diff}[1]{\textcolor{black}{#1}}
\newcommand{\ms}[1]{\textcolor{black}{#1}}

\def\volumeyear{2023}
\graphicspath{{./figs/}{./}} 
\begin{document}

\runninghead{Cossette et. al.}

\title{Decentralized State Estimation: An Approach using Pseudomeasurements and Preintegration}

\author{Charles Champagne Cossette\affilnum{1} and Mohammed Ayman Shalaby\affilnum{1} and David Saussi\'e\affilnum{2} and James Richard Forbes\affilnum{1}}

\affiliation{\affilnum{1}Department of Mechanical Engineering, McGill University\\
\affilnum{2}Department of Electrical Engineering, Polytechnique Montr\'eal}

\corrauth{Charles Champagne Cossette, Department of Mechanical Engineering,
McGill University, Montr\'eal, Canada.}

\email{charles.cossette@mail.mcgill.ca}

\begin{abstract} This paper addresses the problem of decentralized, collaborative state estimation in robotic teams. In particular, this paper considers problems where individual robots estimate similar physical quantities, such as each other's position relative to themselves. The use of \emph{pseudomeasurements} is introduced as a means of modelling such relationships between robots' state estimates, and is shown to be a tractable way to approach the decentralized state estimation problem. Moreover, this formulation easily leads to a general-purpose observability test that simultaneously accounts for measurements that robots collect from their own sensors, as well as the communication structure within the team. Finally, input preintegration is proposed as a communication-efficient way of sharing odometry information between robots, and the entire theory is appropriate for both vector-space and Lie-group state definitions. \ms{To overcome the need for communicating preintegrated-covariance information, a deep autoencoder is proposed that reconstructs the covariance information from the inputs, hence further reducing the communication requirements.} The proposed framework is evaluated on three different simulated problems, and one experiment involving three quadcopters. 
\end{abstract}


\maketitle

\begin{figure*}[h]
  \setlength{\linewidth}{\textwidth}
  \setlength{\hsize}{\textwidth}
  \centering
  \includegraphics[width=0.329\textwidth, clip, trim={5.4cm 4cm 5cm 1.2cm}]{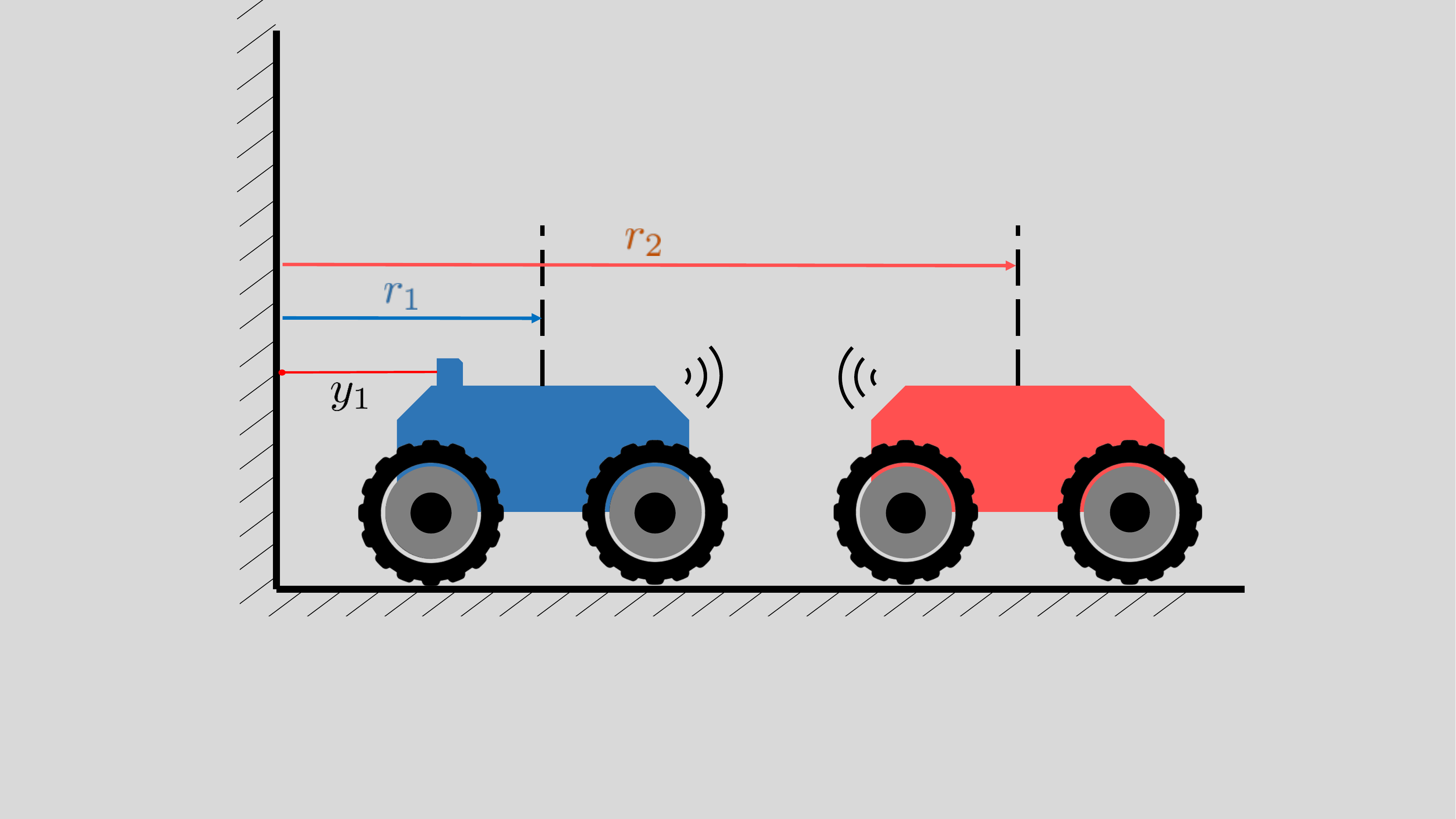}\hspace{0.1cm}%
  \includegraphics[width=0.329\textwidth,
   clip, trim={4cm 5cm 2cm 1.93cm}]{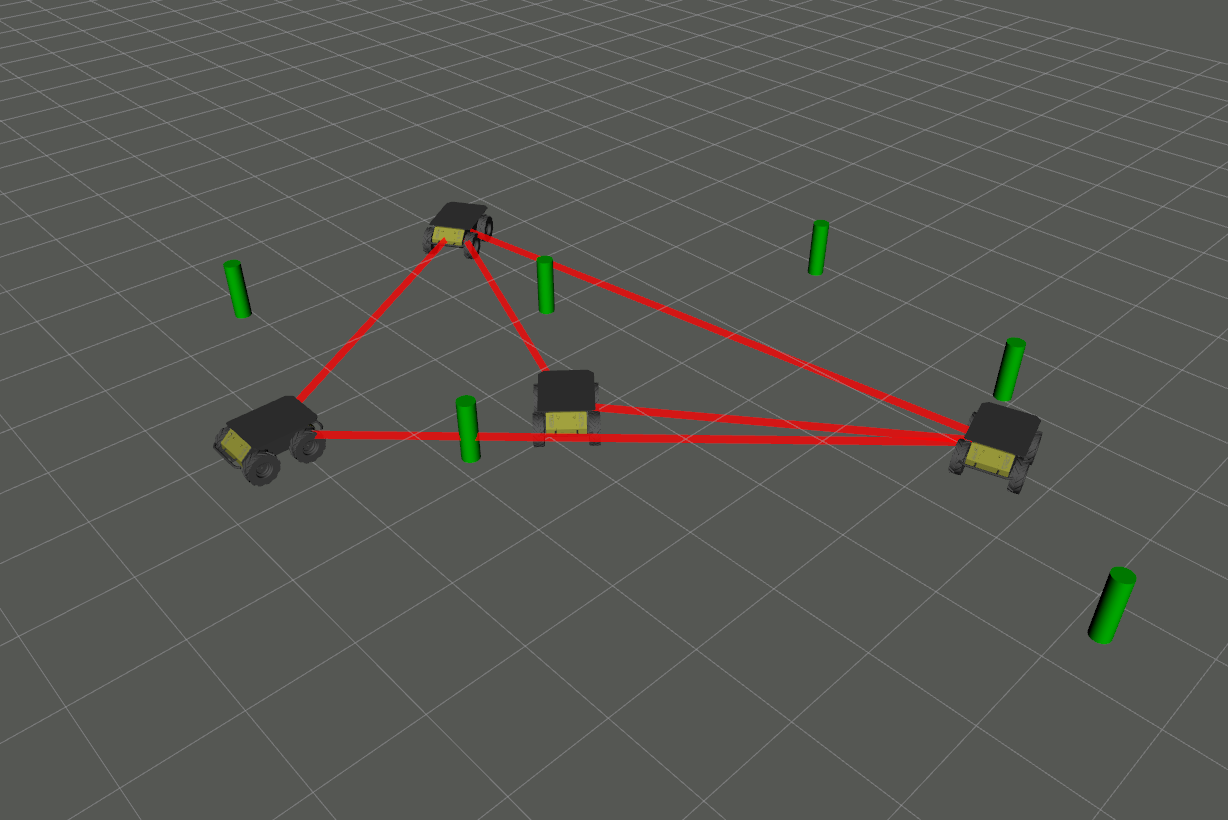}\hspace{0.1cm}%
  \includegraphics[width=0.329\textwidth, clip, trim={0 20.6cm 0 0cm}]{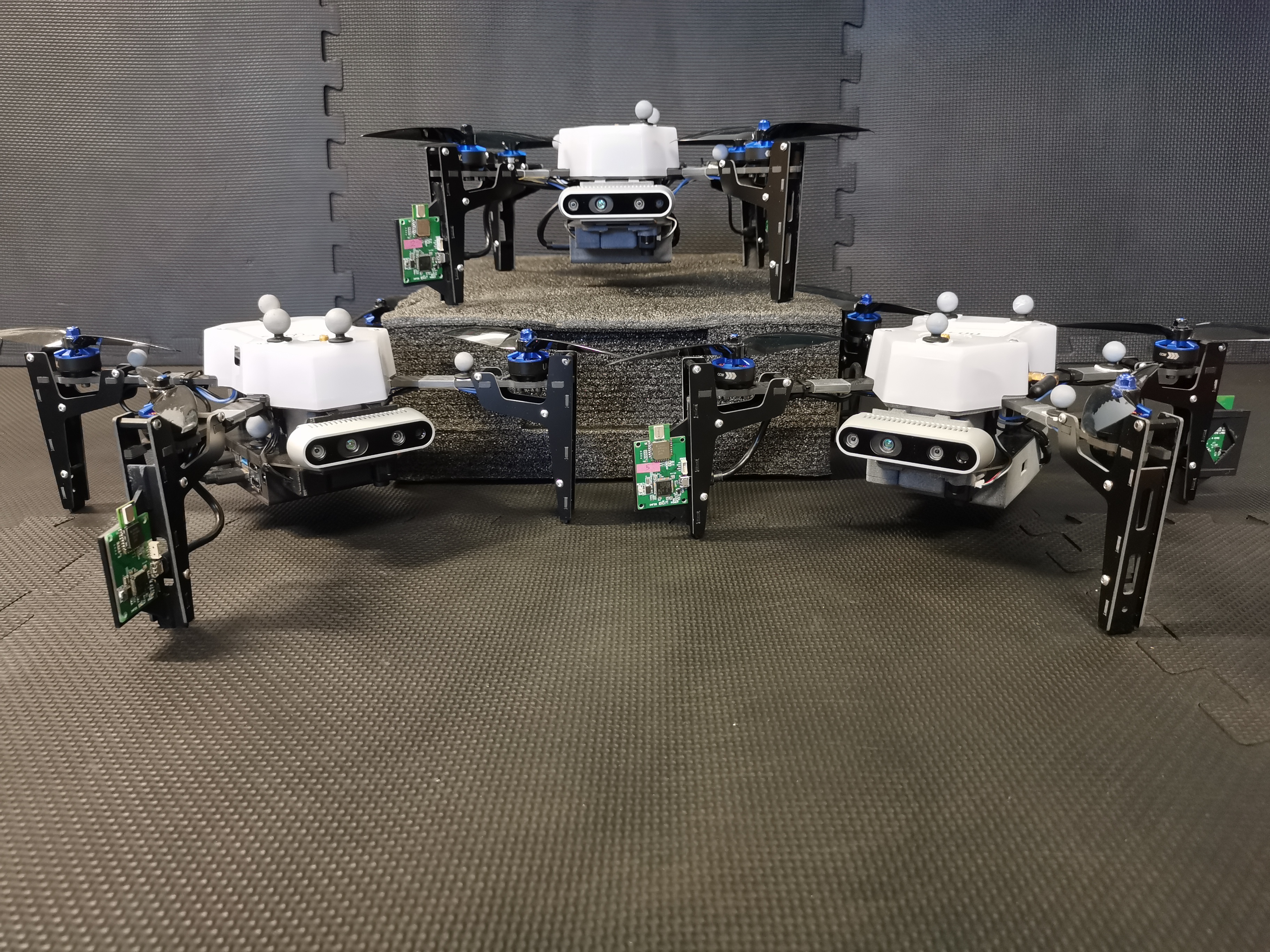}
\caption{Three examples of decentralized estimation problems within the scope of this paper. \textbf{Left:} A toy problem with 1D robots, each estimating both of their positions. \textbf{Middle:} A problem with an incomplete communication graph. Robots observe landmarks, have range measurements to each other, and estimate their own and neighbor absolute poses. \textbf{Right}: A more complicated experimentally-tested problem, where robots equipped with ultra-wideband radios estimate both their own absolute pose and relative poses of neighbors, in addition to IMU biases.}
\label{fig:examples}
\end{figure*}

\section{Introduction}
Decentralized state estimation is a fundamental requirement for real-world multi-robot deployments. Whether the task is collaborative mapping, relative localization, or collaborative dead-reckoning, the multi-robot estimation problem seeks to estimate the state of each robot given \emph{all} the measurements that each robot obtains locally. This problem is made difficult by the fact that not all robots can communicate with each other and, furthermore, that high-frequency sensor measurements would require substantial communication bandwidth to simply share across the team. \diff{A robot might even have insufficient sensors to observe their own state, and hence is dependent on its neighbor's sensors to have a stable estimate. Hypothetically, an estimator that could somehow collect all these sensor measurements on each robot, and fuse them all to jointly estimate the states of every robot in one large system, would have the lowest possible estimation error variance. This is called the \emph{centralized estimator}, but is often infeasible to implement in practice.}

A common approach  is for robots to share their current state and associated covariance rather than of a history of measurement values (\cite{Julier1997, Julier2001, Carrillo-Arce2013}). This approach has the benefit of simplicity, low communication cost, and fixed message size, but suffers from a well-known issue of not being able to compute cross-correlations between the robots' state estimates (\cite{Shalaby2021a}). Furthermore, in certain problems, robots may be estimating the same physical quantities. As an example, consider two robots estimating each other's position, in addition to their own positions, as shown in Figure \ref{fig:examples} (left). Their state vectors are both robots' positions, and therefore both seek to estimate the same physical quantities, a situation referred to here as \emph{full state overlap}. When robot states have similar, if not identical state definitions, it is straightforward to compute the error between their state estimates using simple subtraction. However, for more complicated problems, especially those with state definitions belonging to arbitrary Lie groups, a generalized measure of error between different robots' state estimates must be introduced.

\diff{As a concrete example, which is featured experimentally in this paper, consider the case where quadcopters each possess GPS sensors, IMUs, and inter-robot range measurements using ultra-wideband radio. Suppose one quadcopter loses GPS functionality due to travelling under a bridge or a sensor fault, thus losing absolute positioning information. The challenge is to design an algorithm where this faulty robot maintains accurate absolute positioning estimation by appropriately sharing information with its neighbors.}

\subsection{Contributions}
This paper has three main contributions:
\diff{\begin{enumerate}
    \item a framework for decentralized state estimation that uses \emph{pseudomeasurements} to allow for generic nonlinear relationships between robot states;
    \item a preintegration-based method for constant-time, constant-memory, and constant-communication odometry sharing;
    \item a theory compatible with Lie-group state definitions, including the familiar vector space.
\end{enumerate}}
The pseudomeasurements are shown to be a tractable and effective way to model any generic nonlinear relationship between robot state definitions, including full or partial state overlap as special cases. For example, a nonlinear state relationship is present when robots estimate each other's poses in their own body frames. A pseudomeasurement is introduced for each edge in the communication graph, and the proposed framework also naturally leads to an observability test that takes into account both the local measurements obtained by each robot \emph{and} the communication structure between them. \diff{Usage of pseudomeasurements requires robots to communicate their states and corresponding covariances. Furthermore, the common states between robots must be at the same time step, which potentially requires odometry information to also be shared, so that states can be propagated forward to a common time.} 

The proposed use of preintegration allows sharing odometry information over an arbitrary duration of time, in a lossless matter. The naive alternative is for robots to share a history of odometry measurements since the last time they communicated, which has processing, memory, and communication requirements that grow linearly with the time interval between communications. Preintegration provides a constant-time, constant-memory, and constant-communication alternative that is algebraically identical to simply sharing the input measurements themselves. This makes preintegration a natural choice for multi-robot estimation problems. Moreover, preintegration preserves statistical independence assumptions that typical Kalman filtering prediction steps rely on. Preintegration is best known from the visual-inertial odometry literature (\cite{Lupton2012, Forster2017}), where the same concept, adapted for relative pose estimation is introduced by \cite{Shalaby2023}. This paper generalizes the \diff{multi-robot-preintegration} concept to other common process models in robotics, presents a solution for simultaneous input bias estimation, and also further proposes a deep autoencoder to compress the associated covariance information. 

Finally, the proposed solution is general to any state definition, process model, and measurement model subject to typical Gaussian noise assumptions. \diff{The complexity of the proposed estimation algorithm is identical to a standard extended Kalman filter, and the communicated messages are lightweight and of fixed length. In the experimental test demonstrated in this paper, each robot transmits information at a rate of only 53~kB/s.} 

This paper does not focus on the treatment of cross-correlations, and hence employs the simple, well-known covariance intersection (CI) (\cite{Julier1997, Julier2001}) method. This allows for an arbitrary communication graph within the robot team, while remaining lightweight and avoiding cumbersome bookkeeping. The main drawback is that CI is proven to yield sub-optimal estimation. However, in practice, the performance can remain adequate, and sometimes very comparable to centralized estimation (\cite{Shalaby2021a, Julier2001}), as shown here from simulated and experimental results.

The remainder of this paper is as follows. Related work is discussed in Section~\ref{sec:related_work} and mathematical preliminaries and notation are shown in Section~\ref{sec:prelim}. The paper then starts with a simplified ``toy'' problem showcasing the proposed method in Section~\ref{sec:toy}, and the theory is generalized in Section~\ref{sec:general_problem}. Preintegration sharing is presented in Section~\ref{sec:preintegration}. Finally, Section~\ref{sec:ground_robots} contains an application to a ground robot simulation, Section~\ref{sec:quads} applies the method to a more complicated experimental quadcopter problem.

\section{Related Work} \label{sec:related_work}
There are many sub-problems associated with the overall decentralized state estimation problem. Even if communication links between robots are assumed to be lossless and have infinite bandwidth, there is still the issue of propagating information over general, incomplete communication graphs. For two robots, it is straightforward to compute centralized-equivalent estimators on each robot as done by \cite{Grime1994}, which use information filters to accumulate the information from a series of measurements, and then communicate this quantity. \cite{Grime1994} also derived centralized-equivalent solutions for fully connected graphs, as well as tree-shaped graphs. However, they show that for generic graphs, it is impossible to obtain a centralized-equivalent estimate with only neighboring knowledge, and that more knowledge of the graph topology is required.

\cite{Leung2010} present a general centralized-equivalent algorithm for arbitrary time-varying graphs, and is formulated over distributions directly, hence allowing inference using any algorithm such as an extended or sigma-point Kalman filter. Robots must still share raw measurements with each other, therefore requiring substantial bookkeeping. The approach is extended to the SLAM problem by \cite{Leung2012a}. \cite{Roumeliotis2002} decompose the centralized Kalman filter equations using a singular value decomposition to generate independent equations that each robot can compute. Provided that robots have broadcasting ability, and obtain direct pose measurements of their neighbors, a centralized-equivalent solution can be obtained. The consensus Kalman filter  (\cite{Olfati-Saber2004,Olfati-Saber2005}) aims to asymptotically send the state of $n$ arbitrary nodes to a common value, which is effectively a problem with full state overlap.  \cite{Battistelli2015} proposes alternate consensus approaches such as ``consensus on information'' or ``consensus on measurements''.  However, the problem does not consider the fact that robots collect their own, separate odometry measurements that are not necessarily available to neighboring robots. 

As previously mentioned, one of the simplest solutions to the decentralized estimation problem is to use covariance intersection (\cite{Julier1997, Julier2001}). CI conservatively assumes maximum correlation between robot estimates. Although the performance is theoretically suboptimal, the implementation is extremely simple, and imposes no constraints whatsoever on the communication frequency or graph topology. \cite{Carrillo-Arce2013} apply \ms{CI}
to a collaborative localization problem, where each robot estimates their own absolute state given direct relative pose measurements to other robots. \ms{Meanwhile, \cite{Arambel2001} present a decentralized state estimation algorithm for multiple spacecraft, where each spacecraft estimates the full state of all vehicles and then utilizes CI to fuse the neighbours' full state}. Recently, \ms{split-CI has been introduced to separate states into groups of correlated and independent substates (\cite{Li2013a}),} while \cite{Li2021a} exploit \ms{CI} 
for the fusion of poses on Lie groups. 

\ms{When employing CI, a user-defined weighting parameter has to be chosen, which affects the level of inflation of the block-diagonal components of the covariance matrix. 
 \cite{Zhu2019a} formulate an optimization problem where the logarithm of the determinant of the posterior covariance matrix is minimized as a function of the CI weighting parameter, alongside an alternative linear-matrix-inequality approach that estimates the most conservative posterior covariance matrix. Meanwhile,} \cite{Luft2018} use an EKF-like filter for decentralized estimation where cross-correlations are also explicitly tracked for both the prediction step and the fusion of local measurements. When relative measurements are encountered, an improved approximation to the joint covariance matrix is developed, which outperforms CI. The approach of \cite{Luft2018} assumes that process model inputs between robots are uncorrelated, which is not applicable in some of the problems in this paper. The work by \cite{Jung2020} builds off of \cite{Luft2018} to solve a full 3D collaborative state estimation problem where each robot has a camera and an IMU. 

Another approach using scattering theory has recently been presented for two robots (\cite{Allak2019, Allak2022}), with the objective of reducing the communication cost associated with high-rate sensor measurements. Also making reference to the IMU preintegration technique (\cite{Lupton2012, Forster2017}), covariance pre-computations are derived by \cite{Allak2019} and later extended to also include the mean (\cite{Allak2022}). It is shown that by sharing pre-computed matrices with twice the size as the state vector, a centralized-equivalent state estimate can be directly obtained with no measurement reprocessing. However, the generalization to more robots does not seem straightforward.

\diff{A variety of optimization-based approaches can be seen in the literature, especially when applied to multi-robot simultaneous localization and mapping (SLAM). \cite{Tian2022} have released Kimera-Multi, which uses a distributed pose-graph optimization algorithm to perform metric-semantic SLAM. \cite{Lajoie2023} propose Swarm-SLAM, which performs multi-robot SLAM with an emphasis on using sparsity to minimize the number of data exchanges. However, these distributed SLAM methods are appropriate for situations where each robot has sufficiently rich sensor information via cameras or LIDARs and can perform individual SLAM in the first place. The method of this paper does not impose such a requirement.}

\section{Preliminaries} \label{sec:prelim}
This paper will address problems where an individual robot's process model $\mbf{f}(\cdot)$ and measurement model $\mbf{g}(\cdot)$ are modelled in the standard form of 
\beq
\begin{aligned}
    \mc{X}_{i_k} &= \mbf{f}(\mc{X}_{i_ {k-1}}, \mbf{u}_{i_{k-1}}, \mbf{w}_{i_{k-1}}), \\
    \mbf{y}_{i_k} &= \mbf{g}(\mc{X}_{i_k}) + \mbf{v}_{i_k},
\end{aligned} 
\qquad
\begin{aligned}
   \mbf{w}_{i_{k-1}} &\sim \mc{N}(\mbf{0}, \mbf{Q}_{i_{k-1}}), \\
    \mbf{v}_{i_k} &\sim \mc{N}(\mbf{0},\mbf{R}_{i_k}),
\end{aligned} \label{eq:standard_model}
\eeq
for Robot $i$, where $\mbf{u}_{i_k} \in \mathbb{R}^{n_u}$ is the process input at time step $k$, $\mbf{y}_{i_k} \in \mathbb{R}^{n_y}$ are the measurements, and $\mc{X}_{i_k} \in G$ denotes the robot state belonging to any Lie group $G$. As a notational convenience, the shorthand $\mc{X}_{i:j} = \{\mc{X}_i \; \ldots \; \mc{X}_j \}$ will refer to a collection of arbitrary objects with indices in the range $[i, j]$.

\subsection{Lie groups}
A Lie group $G$ is a smooth manifold whose elements, given a group operation $\circ: G \times G \to G$, satisfy the group axioms (\cite{Sola2018a}). The application of this operation to two arbitrary group elements $\mc{X, Y} \in {G}$ is written as $\mc{X} \circ \mc{Y} \in G$. For any $G$, there exists an associated Lie algebra $\mathfrak{g}$, a vector space identifiable with elements of $\mathbb{R}^m$, where $m$ is referred to as the degrees of freedom of $G$. Lie algebra elements are related to group elements through the exponential and logarithmic maps, denoted $\exp: \mathfrak{g} \to G$ and $\log: G \to \mathfrak{g}$. The ``vee'' and ``wedge'' operators are denoted $(\cdot)^\vee: \mathfrak{g} \to \mathbb{R}^m$ and $(\cdot)^\wedge: \mathbb{R}^m \to \mathfrak{g}$, which can be used to associate Lie algebra elements with vectors. Composing these operators, group elements can be associated with vectors using
\bdis 
\mc{X} = \exp(\mbs{\xi}^\wedge) \triangleq \Exp(\mbs{\xi}), \qquad \mbs{\xi} = \log(\mc{X})^\vee \triangleq   \Log(\mc{X}),
\edis 
where $\mc{X} \in G, \mbs{\xi} \in \mathbb{R}^m$, and the shorthand notation $\Exp: \mathbb{R}^m \to G$ and $\Log:G \to \mathbb{R}^m$ has been defined. Following \cite{Sola2018a}, the adjoint matrix representation of an element $\mc{X}\in G$ is denoted $\mathbf{Ad}:G \to \mathbb{R}^{m \times m}$ and defined such that 
\bdis
\mathbf{Ad}(\mc{X}) \mbs{\xi} = (\mc{X} \mbs{\xi}^\wedge \mc{X}^{-1})^\vee.
\edis
The most common Lie groups appearing in robotics are $SO(n)$, representing rotations in $n$-dimensional space, $SE(n)$, representing poses, and $SE_2(3)$ representing ``extended'' poses that also contain velocity information. In these cases, the elements $\mc{X}$ are invertible matrices and the group operation $\circ$ is regular matrix multiplication.

\subsubsection{$\oplus$ and $\ominus$ operators.}

Estimation theory for vector-space states and Lie groups can be elegantly aggregated into a single mathematical treatment by defining generalized ``addition'' $\oplus: G \times \mathbb{R}^m \to G$ and ``subtraction'' $\ominus:G \times G \to \mathbb{R}^m$ operators, whose precise definitions will depend on the problem at hand. For example, possible implementations include 
\bdis
\begin{array}{ll}
    \mc{X} \oplus \delta \mbf{x} = \mc{X} \circ \Exp(\delta \mbf{x}) & \text{(Lie group right)},\\
    \mc{X} \oplus \delta \mbf{x} =  \Exp(\delta \mbf{x}) \circ  \mc{X} & \text{(Lie group left)},\\
    \mbf{x} \oplus \delta \mbf{x} = \mbf{x} + \delta \mbf{x} & \text{(vector space)},
\end{array}
\edis
for addition and, correspondingly,
\bdis
\begin{array}{ll}
    \mc{X} \ominus \mc{Y} = \Log(\mc{Y}^{-1} \circ \mc{X}) & \text{(Lie group right)},\\
    \mc{X} \ominus \mc{Y} =  \Log(\mc{X} \circ \mc{Y}^{-1})& \text{(Lie group left)},\\
    \mbf{x} \ominus \mbf{y} = \mbf{x} - \mbf{y} & \text{(vector space)},
\end{array}
\edis
for subtraction. This abstraction is natural since a vector space technically qualifies as a Lie group with regular addition $+$ as the group operation. 

\subsubsection{Gaussian distributions on Lie groups.}
As an example use of this abstraction, consider defining a normally-distributed Lie group element with mean $\bar{\mc{X}}$ and covariance $\mbs{\Sigma}$, as done by \cite{Barfoot2014}, with
\bdis 
\mc{X} = \bar{\mc{X}} \circ \Exp(\delta \mbf{x}), \qquad \delta \mbf{x} \sim \mc{N}(\mbf{0}, \mbs{\Sigma}),
\edis
when using a right parameterization, or a similar definition for left parameterizations. This can alternatively be written in an abstract way, applicable to any group or vector space, with
\beq 
\mc{X}  = \bar{\mc{X}} \oplus \delta \mbf{x}, \qquad \delta \mbf{x} \sim \mc{N}(\mbf{0}, \mbs{\Sigma}). \label{eq:gaussian_oplus}
\eeq 
Moreover, given that $\delta \mbf{x} = \mc{X} \ominus \bar{\mc{X}}$, it follows from \eqref{eq:gaussian_oplus} that
\bdis 
p(\mc{X}) = \eta \exp \left(- \frac{1}{2}(\mc{X} \ominus \bar{\mc{X}})^\trans \mbs{\Sigma}^{-1} (\mc{X} \ominus \bar{\mc{X}}) \right)  \triangleq \mc{N}_{L}(\bar{\mc{X}}, \mbs{\Sigma}),
\edis
where the reader should note the definition of the generalized Gaussian $\mc{N}_{L}(\bar{\mc{X}}, \mbs{\Sigma})$ (\cite{Bourmaud2016}).
\subsubsection{Derivatives on Lie groups.}
Again following \cite{Sola2018a}, the Jacobian of a function $f: G \to G$, taken with respect to $\mc{X}$ can be defined as 
\beq
\left.\frac{D f(\mc{X})}{D \mc{X}}\right|_{\bar{\mc{X}} }\triangleq \left.\frac{\p f(\bar{\mc{X}} \oplus \delta \mbf{x}) \ominus f(\bar{\mc{X}})}{\p \delta \mbf{x}}\right|_{\delta \mbf{x} = \mbf{0}}, \label{eq:lie_derivative}
\eeq
where it should be noted that the function  $f(\bar{\mc{X}} \oplus \delta \mbf{x}) \ominus f(\bar{\mc{X}})$ of $\delta \mbf{x}$ has $\mathbb{R}^m$ as both its domain and codomain, and can thus be differentiated using any standard technique. With the above general definition of a derivative, it is easy to define the so-called \emph{ Jacobian of }$G$ as $\mbf{J} = D \Exp(\mbf{x})/ D \mbf{x}$, where left/right group Jacobians are obtained with left/right definitions of $\oplus$ and $\ominus$. 

\subsubsection{Composite groups.} A \emph{composite group} is simply the concatenation of $N$ other Lie groups $G_1, \ldots, G_N$ (\cite{Sola2018a}), with elements of the form 
\bdis 
\mbc{X} = (\mc{X}_1, \ldots \mc{X}_N) \in G_1 \times \ldots \times G_N.
\edis
The group operation, inverse, and identity are defined elementwise. For example,
\beq 
\mbc{X} \circ \mbc{Y} = (\mc{X}_1 \circ \mc{Y}_1 , \ldots, \mc{X}_N \circ \mc{Y}_N).
\eeq
Furthermore, defining $\delta \mbf{x} = [\delta \mbf{x}_1^\trans \; \ldots \; \delta \mbf{x}_N^\trans]^\trans$ the  $\oplus$, operator is given by
\beq
\mbc{X} \oplus \delta \mbf{x} = (\mc{X}_1 \oplus \delta \mbf{x}_1, \; \ldots \;, \mc{X}_N \oplus \delta \mbf{x}_N)
\eeq
and a similar definition applies to $\ominus$.


\subsection{Maximum A Posteriori}
\emph{Maximum a posteriori} (MAP) is the standard approach taken in the robotics literature. Popular algorithms such as the extended Kalman filter (EKF), iterated EKF, sliding-window filter, and batch estimator can all be derived from a MAP approach, thus unifying them under a common theory. Given a statistically independent measurement $\mbf{y}$ with a standard model as in \eqref{eq:standard_model}, as well as a prior distribution $p(\mc{X}) = \mc{N}_L(\check{\mc{X}}, \mbfcheck{P})$, the estimate $\mchat{X}$ produced by the MAP approach is 
\begin{align}
\hat{\mc{X}} &= \argmax_{\mc{X}} p(\mc{X} | \mbf{y})\\
&= \argmax_{\mc{X}} \eta \;p(\mbf{y} | \mc{X}) p(\mc{X})\\
&= \argmax_{\mc{X}} \eta \;\mc{N}(\mbf{g}(\mc{X}), \mbf{R})\; \mc{N}_L(\check{\mc{X}}, \mbfcheck{P})
\end{align}
where $\eta$ is a normalization constant. Equivalently, minimizing the negative logarithm yields a nonlinear least-squares problem of the form 
\begin{align} 
\hat{\mc{X}} &= \argmax_{\mc{X}} \frac{1}{2} \mbf{e}(\mc{X})^\trans \mbf{W} \mbf{e}(\mc{X}),\label{eq:map_ls} \\
\mbf{e}(\mc{X}) &= \bma{c} \mc{X} \ominus \check{\mc{X}} \\ \mbf{y} - \mbf{g}(\mc{X}) \ema, 
\end{align}
where $\mbf{W} = \mathrm{diag}(\mbfcheck{P}^{-1}, \mbf{R}^{-1})$. Using an on-manifold optimization approach, \eqref{eq:map_ls} can be solved by first parameterizing the state with $\mc{X} = \hat{\mc{X}} \oplus \delta \mbf{x}$ and solving the problem 
\beq 
{\delta \mbfhat{x}} = \argmax_{\delta \mbf{x}} \frac{1}{2} \mbf{e}(\hat{\mc{X}} \oplus \delta \mbf{x})^\trans \mbf{W} \mbf{e}(\hat{\mc{X}} \oplus \delta \mbf{x}). \label{eq:map_perturbed}
\eeq 
Using \eqref{eq:lie_derivative}, the Jacobian of the error vector is given by
\beq 
\mbf{H} \triangleq \left.\frac{D \mbf{e}(\mc{X})}{D \mc{X}}\right|_{\bar{\mc{X}}}=  \left.\frac{\p \mbf{e}(\hat{\mc{X}} \oplus \delta \mbf{x})}{\p \delta \mbf{x}}\right|_{\delta \mbf{x} = \mbf{0}},
\eeq
and an approximate solution to \eqref{eq:map_perturbed} can be obtained by solving the Gauss-Newton system 
\beq 
(\mbf{H}^\trans \mbf{W} \mbf{H})\delta \mbfhat{x} = \mbf{H}^\trans \mbf{W} \mbf{e}(\hat{\mc{X}}).
\eeq
The above is iterated with $\hat{\mc{X}} \gets \hat{\mc{X}} \oplus \delta \mbfhat{x}$ and initialized with $\hat{\mc{X}} \gets \check{\mc{X}}$. A common approximation for the posterior covariance $\mbfhat{P}$ where $p(\mc{X}|\mbf{y}) \approx \mc{N}_L(\hat{\mc{X}}, \mbfhat{P})$  is given by $\mbfhat{P} = (\mbf{H}^\trans \mbf{W} \mbf{H})^{-1}$ with $\mbf{H}$ evaluated at $\mchat{X}$. 
\subsection{Covariance Intersection}
Covariance intersection (CI) is a tool introduced by \cite{Julier1997} for the purposes of decentralized data fusion under unknown cross-correlations, and can be summarized with the following \diff{lemma}.
\begin{lemma}[\diff{Consistency of Covariance Intersection}] \label{thm:ci}
    The inequality 
    \beq 
    \bma{cc} \frac{1}{w}\mbs{\Sigma}_{xx} & \mbf{0} \\ \mbf{0} & \frac{1}{1-w}\mbs{\Sigma}_{yy} \ema \geq \bma{cc} \mbs{\Sigma}_{xx}& \mbs{\Sigma}_{xy} \\ \mbs{\Sigma}_{xy}^\trans & \mbs{\Sigma}_{yy}\ema,\label{eq:ci_consistency}
    \eeq
    which applies in the positive definite sense, holds for all $w \in (0,1)$, where $\mbs{\Sigma}_{xx}, \mbs{\Sigma}_{yy},$ and the right-hand-side of \eqref{eq:ci_consistency} are positive definite.
\end{lemma}
\diff{There are several known strategies for choosing $w$ (\cite{Julier2001}). Following \cite{Shalaby2021a}, a fixed value of $w=0.99$ is chosen for all the results shown in this paper, as it is a simple approach that yields acceptable results.}

\section{A Toy Problem} \label{sec:toy}
Consider first one of the simplest multi-robot estimation problems, shown on the left of Figure \ref{fig:examples}. Two robots are located at positions $r_1$ and $r_2$, respectively, and both robots seek to estimate both robots' positions. By design, each robot carries distinct, conceptually independent estimates, even though their states represent the same true \emph{physical} variables. This mimics exactly what will occur in implementation, as each robot's processor will have a live estimate of both robots' positions. Their state vectors can therefore be defined as 
\beq 
\mbf{x}_1 = \bma{cc}r_1^{[1]}&r_2^{[1]}\ema^\trans, \qquad
\mbf{x}_2 = \bma{cc}r_1^{[2]}& r_2^{[2]}\ema^\trans,
\eeq
where the square bracket superscript $(\cdot)^{[i]}$ is used when necessary to denote Robot $i$'s estimate or ``instance'' of a common physical variable. Each robot also collects local measurements from its sensors. Robot 1 is capable of measuring its own position, 
\beq 
{y}_1 = \mbf{G}_1\mbf{x}_1 + {v}_1, \quad {v}_1 \sim \mc{N}({0}, {R}_1), \quad \mbf{G}_1 = \bma{cc}1 & 0\ema,
\eeq 
while Robot 2 is only capable of measuring its position relative to Robot 1,
\beq 
{y}_2 = \mbf{G}_2 \mbf{x}_2 + {v}_2, \quad {v}_2 \sim \mc{N}({0}, {R}_2), \quad \mbf{G}_2 = \bma{cc}-1& 1\ema.
\eeq
To keep things simple for this demonstrative problem, robots are assumed to have access to each other's input measurements, such as wheel odometry. This allows them to predict their state forward in time using a conventional Kalman filter. However, a more communication-efficient solution will be proposed in Section \ref{sec:preintegration}. Neither robot is capable of estimating their full state vector from local measurements only, meaning that some form of communication will be required. 

To reflect the knowledge that the two robots' state vectors are physically the same, a key design choice of this paper is to incorporate a \emph{pseudomeasurement} of the form 
\beq 
\mbf{y}_{12} = \mbf{x}_1 - \mbf{x}_2 + \mbf{v}_{12}, \qquad \mbf{v}_{12} \sim \mc{N}(\mbf{0}, \mbs{\Psi}),
\eeq
whose ``measured'' value is always exactly zero. This pseudomeasurement can be viewed as a soft constraint on the problem, inversely weighted by the \diff{arbitrary} pseudomeasurement covariance $\mbs{\Psi}$. The estimation problem is now to compute, as accurately as possible, the posterior distribution
\beq 
p(\mbf{x}_1, \mbf{x}_2 | {y}_1, {y}_2, \mbf{y}_{12}).
\eeq

\subsection{Solution via MAP}
Applying MAP to this simplified problem is to say that 
\beq 
\mbfhat{x}_1, \mbfhat{x}_2 = \argmax_{\mbf{x}_1, \mbf{x}_2} 
p(\mbf{x}_1, \mbf{x}_2 | {y}_1, {y}_2, \mbf{y}_{12}). \label{eq:map}
\eeq
Assuming that ${v}_1, {v}_2, \mbf{v}_{12}$ are all independent random variables, \diff{i.e., that $p(v_1, v_2, \mbf{v}_{12}) = p(v_1)p(v_2)p(\mbf{v}_{12})$,}  allows the use of Bayes' rule to write 
\begin{multline}
p(\mbf{x}_1, \mbf{x}_2 | {y}_1, {y}_2, \mbf{y}_{12}) = \eta p({y}_1 | \mbf{x}_1)p({y}_2 | \mbf{x}_2)\\
\times p(\mbf{y}_{12}|\mbf{x}_1, \mbf{x}_2)p(\mbf{x}_1, \mbf{x}_2),
\label{eq:map2}
\end{multline}
\diff{where $\eta$ is a normalization constant that does not depend on $\mbf{x}_1$ or $\mbf{x}_2$. } Next, assume that the prior distributions of the robots are independent and Gaussian, possibly as a result of using CI,
\beq 
p(\mbf{x}_1, \mbf{x}_2) = p(\mbf{x}_1)p(\mbf{x}_2) = \mc{N}(\mbfcheck{x}_1, \mbfcheck{P}_1)\mc{N}(\mbfcheck{x}_2, \mbfcheck{P}_2).\label{eq:map3}
\eeq
Substituting \eqref{eq:map3} into \eqref{eq:map2} and grouping terms into those available to each robot yields 
\begin{multline}
p(\mbf{x}_1, \mbf{x}_2 | {y}_1, {y}_2, \mbf{y}_{12}) = \\ \eta p(\mbf{y}_{12}|\mbf{x}_1, \mbf{x}_2) \Big(p({y}_1 | \mbf{x}_1) p(\mbf{x}_1) \Big) \Big(p({y}_2 | \mbf{x}_2)p(\mbf{x}_2) \Big).
\label{eq:map4}
\end{multline}
Since the local measurement models are linear, it is straightforward to exactly compute the terms
\beq 
p({y}_i | \mbf{x}_i) p(\mbf{x}_i) = \eta_i p(\mbf{x}_i|{y}_i) = \eta_i \mc{N}(\mbftilde{x}_i, \mbftilde{P}_i), \quad i = 1,2, \label{eq:map5}
\eeq
using the regular Kalman filter equations. The means and covariances $\mbftilde{x}_i, \mbftilde{P}_i$ (with tildes) represent the distribution of each robot's state conditioned on only local measurements, without the information that the robots' states are physically the same. Substituting \eqref{eq:map5} into \eqref{eq:map4} yields a simplified expression for the posterior, and the  optimization problem \eqref{eq:map} now leads to the least-squares problem
\beq 
\mbfhat{x}_1, \mbfhat{x}_2 = \argmin_{\mbf{x}_1, \mbf{x}_2}\frac{1}{2}\mbf{e}(\mbf{x}_1, \mbf{x}_2)^\trans \mbf{W}\mbf{e}(\mbf{x}_1, \mbf{x}_2),
\eeq
where 
\begin{align*}
    \mbf{e}(\mbf{x}_1, \mbf{x}_2) &= \bma{cc} \mbf{1} & \mbf{0} \\ \mbf{0} & \mbf{1} \\ \mbf{1} & -\mbf{1} \ema \bma{c} \mbf{x}_1 \\ \mbf{x}_2 \ema  - \bma{c}\mbftilde{x}_1 \\ \mbftilde{x}_2 \\ \mbf{0} \ema \triangleq \mbf{H}\mbf{x} - \mbf{z}, \\
    \mbf{W} &= \mathrm{diag}(\mbftilde{P}_1^{-1},\mbftilde{P}_2^{-1}, \mbs{\Psi}^{-1}).
\end{align*}
In this linear case the unique solution $\mbfhat{x}$ is given by 
\beq 
\bma{c} \mbfhat{x}_1 \\ \mbfhat{x}_2 \ema = (\mbf{H}^\trans \mbf{W} \mbf{H})^{-1}\mbf{H}^\trans \mbf{W} \mbf{z}, \label{eq:map_system}
\eeq
which is also known to be the mean. The relevant matrices expand to 
\begin{align}
    \mbf{H}^\trans \mbf{W} \mbf{H} &= \bma{cc} \mbftilde{P}_1^{-1} + \mbs{\Psi}^{-1} & - \mbs{\Psi}^{-1} \\ -\mbs{\Psi}^{-1} & \mbftilde{P}_2^{-1} + \mbs{\Psi}^{-1}\ema, \\
    \mbf{H}^\trans \mbf{W} \mbf{z} &= \bma{c} \mbftilde{P}_1^{-1} \mbftilde{x}_1 \\  \mbftilde{P}_2^{-1} \mbftilde{x}_2 \ema.
\end{align}
\diff{The next steps involve various applications of the Sherman-Morrison-Woodbury (SMW) identities to analytically invert the inverse covariance matrix $\mbf{H}^\trans \mbf{W} \mbf{H}$, as well as solve for the solution using \eqref{eq:map_system}. The derivation details are omitted for brevity but follow the same steps as \cite[Ch.~3.3.2]{Barfoot2022}. }The eventual result is
\begin{align*}
    \mbfhat{x} &\triangleq \bma{c} \mbfhat{x}_1 \\ \mbfhat{x}_2 \ema =  \bma{c}\mbftilde{x}_1 + \mbf{K}_1(\mbftilde{x}_2 - \mbftilde{x}_1) \\ \mbftilde{x}_2 + \mbf{K}_2(\mbftilde{x}_1 - \mbftilde{x}_2) \ema ,\\ 
    \mbfhat{P} &\triangleq (\mbf{H}^\trans \mbf{W} \mbf{H})^{-1} \\ 
    &= \bma{cc} (\mbf{1} - \mbf{K}_1)\mbftilde{P}_1 & -\mbf{K}_1\mbftilde{P}_2 \\ -\mbf{K}_2\mbftilde{P}_1 & (\mbf{1} - \mbf{K}_2) \mbftilde{P}_2 \ema , \numberthis \label{eq:toy_covariance}\\
    \mbf{K}_1 &\triangleq \mbftilde{P}_1(\mbs{\Psi} + \mbftilde{P}_2 + \mbftilde{P}_1)^{-1}, \\
    \mbf{K}_2 & \triangleq \mbftilde{P}_2 (\mbs{\Psi} + \mbftilde{P}_2 + \mbftilde{P}_1)^{-1}, 
\end{align*}
and furthermore $p(\mbf{x}_1, \mbf{x}_2|\mbf{y}_1 , \mbf{y}_2, \mbf{y}_{12}) = \mc{N}(\mbfhat{x}, \mbfhat{P})$, \diff{which is a standard result from MAP approaches (\cite{Barfoot2022}).} The final individual estimates are obtained by marginalizing out the other robots' states, which is trivial to do in covariance form by simply extracting the corresponding blocks from $\mbfhat{x}, \mbfhat{P}$, yielding
\begin{align}
    p(\mbf{x}_1| \mbf{y}_1, \mbf{y}_2, \mbf{y}_{12}) &= \mc{N}(\mbfhat{x}_1, (\mbf{1} - \mbf{K}_1)\mbftilde{P}_1), \\
    p(\mbf{x}_2| \mbf{y}_1, \mbf{y}_2, \mbf{y}_{12}) &= \mc{N}(\mbfhat{x}_2, (\mbf{1} - \mbf{K}_2)\mbftilde{P}_2).
\end{align} 
\diff{The equations \eqref{eq:toy_covariance} have a form similar to a situation where robots simply treated the other robot's state estimate as a ``measurement'' of their own state. This is a result that is specific to this simple toy problem, where robots have full state overlap.}

Conditioning on the pseudomeasurement $\mbf{y}_{12}$ has introduced cross-correlation terms in \eqref{eq:toy_covariance}, which are feasible to keep track of for this two-robot scenario, but introduce substantial complexity for an arbitrary multi-robot scenario. Therefore, this paper simply employs the CI approximation as required, \diff{including for this toy problem for the sake of consistency. Specifically, before each state fusion using \eqref{eq:toy_covariance}, inflate the covariance matrices with 
\beq 
 \mbftilde{P}_1 \gets \frac{1}{w} \mbftilde{P}_1, \qquad \mbftilde{P}_2 \gets \frac{1}{1-w} \mbftilde{P}_2, \label{eq:toy_ci}
\eeq
where $w=0.99$ is used.} 

\begin{figure}[t]
    \centering
    \includegraphics[width=\linewidth, clip, trim={0 0.3cm 0 0}]{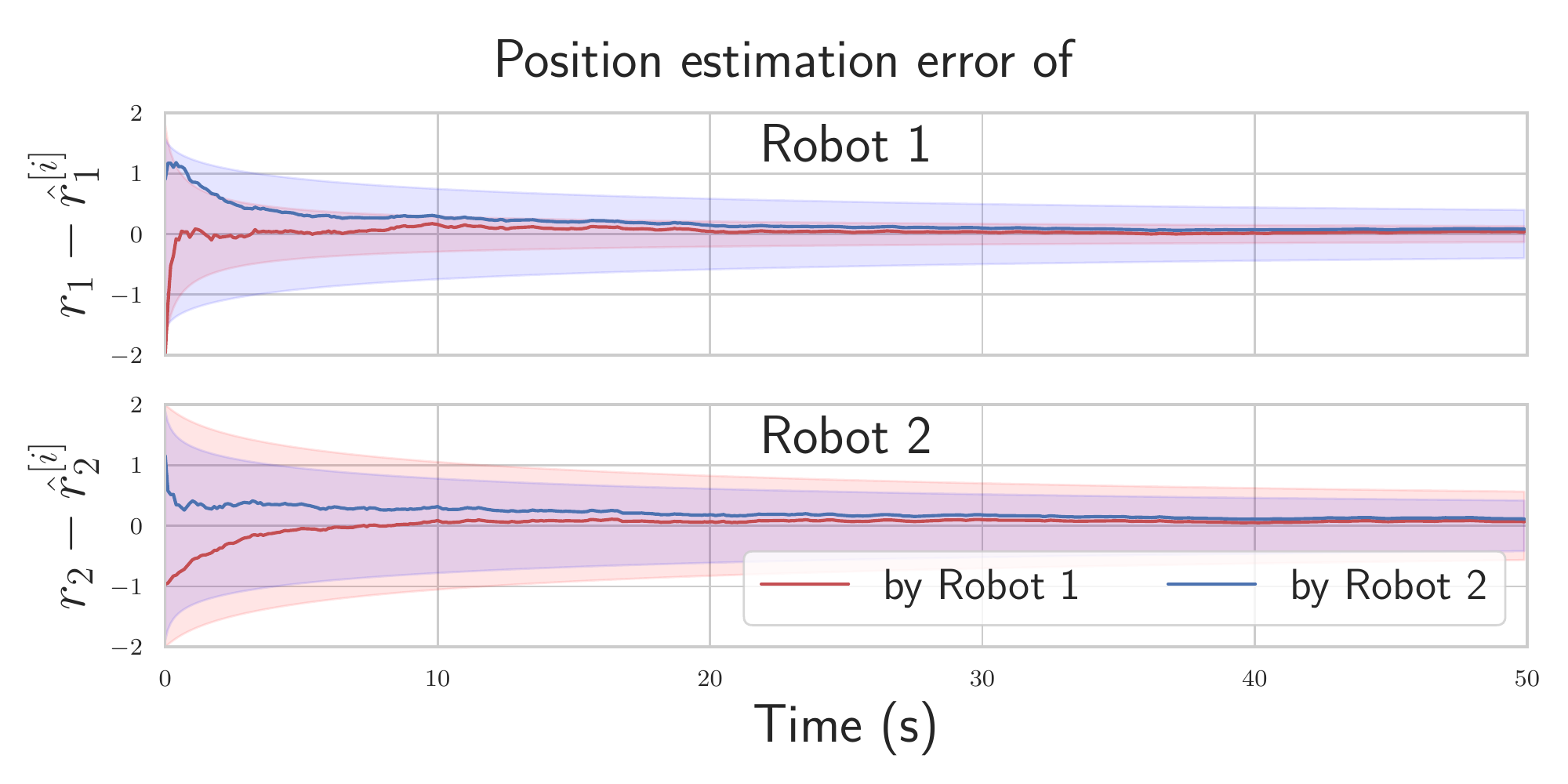}
    \caption{Estimation convergence for a single trial of the two-robot toy problem with $\mbs{\Psi} = 10 \cdot \mbf{1}$. Due to pseudomeasurements, the robot states successfully converge to a common value.}
    \label{fig:toy_convergence}
\end{figure}

Figure \ref{fig:toy_convergence} shows the estimation error of each robot as multiple pseudomeasurements are fused in succession. The two robots' estimates not only converge to zero error, but also to a common value, which is the main effect of the pseudomeasurent. \diff{A pseudomeasurement covariance of $\mbs{\Psi} = 10 \cdot \mbf{1}$ was chosen for this simulation to show its effect, but smaller values can be used. Since the expressions in \eqref{eq:toy_covariance} are in covariance form, it is even possible to use $\mbs{\Psi} = \mbf{0}$, in which case the two estimates will converge together after the first pseudomeasurement. For the prior distributions, arbitrary Gaussian distributions were chosen, with the initial true states drawn from these distributions.}

\begin{figure}[t]
    \centering
    \includegraphics[width=\linewidth, clip, trim={0 1cm 0 0}]{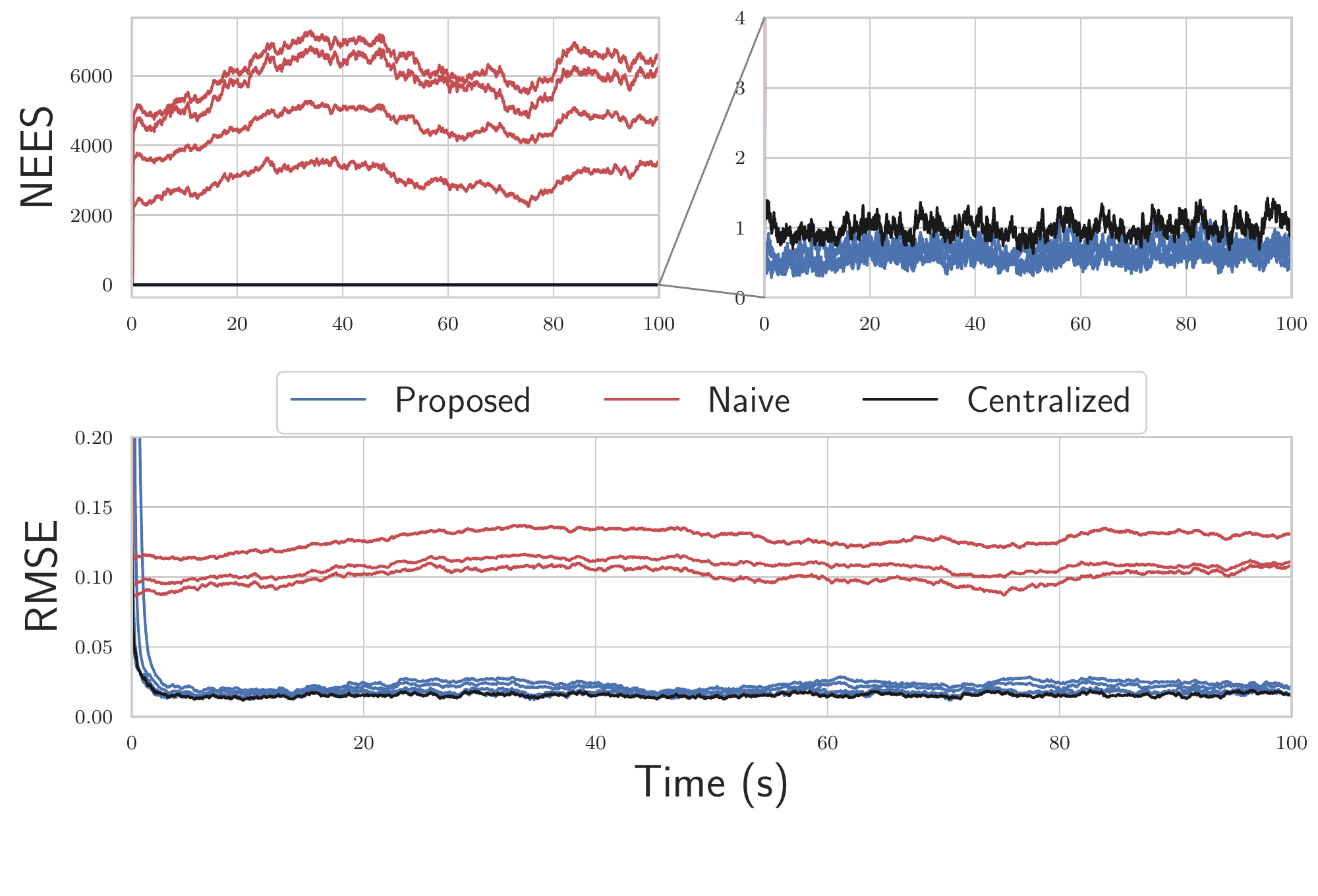}
    \caption{Results of 100 Monte Carlo trials for a four-robot version of the toy problem. The top two plots consist of a NEES plot, which is a measure of consistency. The bottom plot is the RMSE of the state. The proposed solution, which performs CI, remains statistically consistent and has reasonably low error in many cases.}
    \label{fig:toy_monte_carlo}
\end{figure}

In Figure \ref{fig:toy_monte_carlo}, 100 Monte-Carlo trials are performed on a simulation of this toy problem, but extended to four robots \diff{using the methods from the next section}. The root-mean-squared error (RMSE) and normalized estimation error squared (NEES), calculated as per \cite[Ch.~5.4]{Bar-Shalom2001}, are plotted through time. The lines marked ``Proposed'' fuse pseudomeasurements as described, and use CI before each state fusion. The naive solution is identical, \diff{but does not perform the covariance intersection step in \eqref{eq:toy_ci}} before state fusion, thus completely neglects cross-correlations. \diff{The centralized solution is simply a Kalman filter with state $\mbf{x} = [r_1\;\;r_2]^T$ fusing both the measurements $y_1$ and $y_2$ using the standard equations. } Although the use of CI does introduce error compared to the centralized solution, it is still vastly better than the naive approach.

\section{General Problem} \label{sec:general_problem}
Consider now $N$ robots, which communicate in correspondance with an arbitrary undirected graph $\mc{G} = (\mc{V}, \mc{E})$ where 
$\mc{V} = \{1, \ldots, N\}$ is the set of nodes or robot IDs, and $\mc{E}$ is the set of edges. The robots have states $\mc{X}_i \in G_i, \; i \in \mc{V}$ belonging to possibly different groups. Pseudomeasurement models $\mbf{c}_{ij}: G_i \times G_j \to \mathbb{R}^c$ are now defined in a generic way
\beq 
\mbf{y}_{ij_{k}} = \mbf{c}_{ij}(\mc{X}_{i_{k}}, \mc{X}_{j_{k}}) + \mbf{v}_{ij_k}, \qquad \mbf{v}_{ij_k} \sim \mc{N}(\mbf{0}, \mbs{\Psi}),
\eeq
for each pair $(i, j) \in \mc{E}$.  \diff{These models are designed by the user, and should correspond with redundant, common, or over-parameterizations of states appearing on different robots, such as when two robots are estimating the same physical quantities. However, these quantities can also differ in the way they are represented from robot to robot, such as each robot resolving the same physical pose in their own frame, and hence the pseudomeasurement function is left as a general nonlinear function. Examples shall be given in Section \ref{sec:ground_robots} and \ref{sec:quads}.}

Using MAP, the general state fusion problem is now 
\beq 
\hat{\mc{X}}_{1:N} = \argmax_{\mc{X}_{1:N}} p(\mc{X}_{1:N}|\mbf{y}_{ij}), \;\;\text{  for all  } (i,j) \in \mc{E}.
\eeq
It is easier to instead consider a single pseudomeasurement at a time, such as, without loss of generality, $\mbf{y}_{12}$. The posterior distribution given only $\mbf{y}_{12}$ is
\begin{align}
p(\mc{X}_{1:N}|\mbf{y}_{12}) &= \eta p(\mbf{y}_{12}| \mc{X}_1, \mc{X}_2)p(\mc{X}_{1:N}) \\
&= \eta \; \mc{N}(\mbf{c}_{12}(\mc{X}_{1}, \mc{X}_{2}), \mbs{\Psi}) \; \prod_{i=1}^N \mc{N}_{L}(\tilde{\mc{X}}_i, \mbftilde{P}_i)\quad \label{eq:single_pseudo}
\end{align}
where robot state priors have again been assumed to be independent, as a result of using covariance intersection. Due to this independence, the variables $\mc{X}_{3:N}$ can be removed from the optimization problem since their optimal values are simply $\mchat{X}_{3:N} =\tilde{\mc{X}}_{3:N}$ and have no effect on $\mc{X}_1, \mc{X}_2$. Minimizing the negative logarithm of  \eqref{eq:single_pseudo}, omitting terms involving $\mc{X}_{3:N}$, leads to a least-squares problem with error vector  given by 
\beq 
\mbf{e}(\mc{X}_{1}, \mc{X}_2) = \bma{c} \mc{X}_1 \ominus \tilde{\mc{X}}_1 \\ \mc{X}_{2} \ominus \tilde{\mc{X}}_2 \\ -\mbf{c}_{12}(\mc{X}_{1}, \mc{X}_{2}) \ema,
\eeq
and weight $ \mbf{W} = \mathrm{diag}(\mbftilde{P}_1^{-1}, \mbftilde{P}_2^{-1}, \mbs{\Psi}^{-1})$. Defining $\mbf{J}_i$ as the group Jacobian associated with $G_i$, as well as $\mbf{S}_i, \mbf{S}_j$ being the Jacobians of $\mbf{c}_{ij}$ with respect to $\mc{X}_i, \mc{X}_j$, respectively, the Jacobian of the error vector is 
\beq 
\mbf{H} = \bma{cc} \mbf{J}_1^{-1} & \mbf{0}   \\  \mbf{0} & \mbf{J}_2^{-1}  \\ -\mbf{S}_1 & -\mbf{S}_{2}  \ema,
\eeq
which is written without arguments $(\mc{X}_1, \mc{X}_2)$ for brevity. The relevant terms of the Gauss-Newton system are 
\begin{multline*}
\mbf{H}^\trans \mbf{W} \mbf{H} = \\\bma{cc}\mbf{J}_1^{-\trans} \mbf{P}_1^{-1} \mbf{J}_1^{-1} + \mbf{S}_1^\trans \mbs{\Psi}^{-1} \mbf{S}_1 & \mbf{S}_1^\trans \mbs{\Psi}^{-1} \mbf{S}_2 \\ \mbf{S}_2^\trans \mbs{\Psi}^{-1} \mbf{S}_1 & \mbf{J}_2^{-\trans} \mbf{P}_2^{-1} \mbf{J}_2^{-1} + \mbf{S}_2^\trans \mbs{\Psi}^{-1} \mbf{S}_2 \ema,
\end{multline*}
\vspace{-12pt}
\begin{multline*}
    \mbf{H}^\trans \mbf{W} \mbf{e}(\mc{X}_1, \mc{X}_2)= \\
    \bma{c}\mbf{J}_1^{-\trans} \mbf{P}_1^{-1}(\mc{X}_1 \ominus \tilde{\mc{X}_1}) - \mbf{S}_1^\trans \mbs{\Psi}^{-1} \mbf{c}_{12}(\mc{X}_1, \mc{X}_2) \\\mbf{J}_2^{-\trans} \mbf{P}_2^{-1}(\mc{X}_2 \ominus \tilde{\mc{X}_2}) - \mbf{S}_2^\trans \mbs{\Psi}^{-1} \mbf{c}_{12}(\mc{X}_1, \mc{X}_2) \ema ,
\end{multline*}
which, by substantial manipulation with the SMW identities, can be used to analytically compute $\delta \mbfhat{x} = (\mbf{H}^\trans \mbf{W} \mbf{H})^{-1}\mbf{H}^\trans \mbf{W} \mbf{e}(\hat{\mc{X}}_1, \hat{\mc{X}}_2)$, producing on-manifold iterated-EKF-like expressions. The result is 
\beq
\begin{aligned}
    \delta \mbfhat{x}_1 &= - \mbf{J}_1(\hat{\mc{X}}_1 \ominus \mctilde{X}_1) + \mbf{K}_1 \mbf{z}, \\ 
    \delta \mbfhat{x}_2&= - \mbf{J}_2(\hat{\mc{X}}_2 \ominus \mctilde{X}_2) + \mbf{K}_2 \mbf{z}, \\
    \mbf{K}_1 &= \mbf{J}_1 \mbftilde{P}_1 \mbf{J}_1^\trans \mbf{S}_1^\trans \mbf{V}^{-1},\\
    \mbf{K}_2 &= \mbf{J}_2 \mbftilde{P}_2 \mbf{J}_2^\trans \mbf{S}_2^\trans \mbf{V}^{-1},\\
    \mbf{z} &= -\mbf{c}_{12}(\mctilde{X}_1, \mctilde{X}_2) + \mbf{S}_1 \mbf{J}_1(\mchat{X}_1 \ominus \mctilde{X}_1)\\
    &\qquad + \mbf{S}_2 \mbf{J}_2(\mchat{X}_2 \ominus \mctilde{X}_2) ,\\
    \mbf{V} &= \mbs{\Psi} + \mbf{S}_1\mbf{J}_1 \mbftilde{P}_1 \mbf{J}_1^\trans \mbf{S}_1^\trans+ \mbf{S}_2\mbf{J}_2 \mbftilde{P}_2 \mbf{J}_2^\trans \mbf{S}_2^\trans, \label{eq:general_mean}
\end{aligned}
\eeq
where iteration is done with $\mchat{X}_i \gets \mchat{X}_i \oplus \delta \mbfhat{x}_i$, after initialization with $\mchat{X}_i \gets \mctilde{X}_i$, \diff{until a convergence condition is met, such as $\delta \mbfhat{x}_i$ being sufficiently small.} The marginal posterior covariances of Robots 1 and 2 are obtained from the corresponding diagonal blocks of $(\mbf{H}^\trans \mbf{W} \mbf{H})^{-1}$, 
and can be shown to be 
\beq
\begin{aligned}
    \mbfhat{P}_1 &= (\mbf{1} - \mbf{K}_1 \mbf{S}_1) \mbf{J}_1 \mbftilde{P}_1 \mbf{J}_1^\trans, \\
    \mbfhat{P}_2 &= (\mbf{1} - \mbf{K}_2 \mbf{S}_2) \mbf{J}_2 \mbftilde{P}_2 \mbf{J}_2^\trans. \label{eq:general_cov}
\end{aligned}
\eeq
The above fusion step introduces cross-correlations between the state estimates of $\mc{X}_1$ and $\mc{X}_2$, and hence require a covariance intersection step 
\beq
    \mbfhat{P}_1 \gets \frac{1}{w} \mbfhat{P}_1, \qquad \mbfhat{P}_2 \gets \frac{1}{1-w} \mbfhat{P}_2, \qquad w \in (0,1).
\eeq
The next step is to make the approximation that $p(\mc{X}_{1:N}|\mbf{y}_{12}) \approx \prod_{i=1}^N \mc{N}_L(\mchat{X}_i, \mbfhat{P}_i)$, and proceed with the fusion of a second pseudomeasurement, using $p(\mc{X}_{1:N}| \mbf{y}_{12}, \mbf{y}_{13}) = \eta p(\mbf{y}_{13}|\mc{X}_1, \mc{X}_3) p(\mc{X}_{1:N}|\mbf{y}_{12})$. This new posterior can again be approximated as Gaussian using the expressions in \eqref{eq:general_mean} and \eqref{eq:general_cov}, and the process is repeated until all pseudomeasurements are incorporated. \diff{The order in which the pseudomeasurements are fused is arbitrary and will correspond to the times at which information is shared between robots.} 

\diff{If only one iteration is performed, equations \eqref{eq:general_mean} and \eqref{eq:general_cov} simplify to the on-manifold EKF equations since $\mbf{J}_i(\mctilde{X}_i \ominus \mctilde{X}_i) = \mbf{J}_i(\mbf{0}) = \mbf{1}$. Much like the EKF is often sufficient compared to the iterated EKF, performing a single iteration in this multi-robot case is also sufficient for some problems.}  Algorithm~\ref{alg:alg1} summarizes the general-purpose decentralized estimation algorithm from the point of view of an arbitrary robot. The algorithm is presented in a callback format, describing how the current state estimate is updated when various events occur.


\begin{algorithm}[H]
    \caption{Decentralized estimation with no input sharing.}\label{alg:alg1}
    For Robot $i$ with process and measurement models given by \eqref{eq:standard_model}, the following points break down how to compute the estimate when various events occur. Robot $i$'s estimate is initialized with $\mccheck{X}_{i_0}, \mbfcheck{P}_{i_0}$. The matrices $\mbf{F}_{i_k}, \mbf{L}_{i_k}$ are the Jacobians of $\mbf{f}$ with respect to $\mc{X}_{i_k}$ and $\mbf{w}_{i_k}$, respectively. The matrix $\mbf{G}_{i_k}$ is the Jacobian of $\mbf{g}$, and $\mbf{S}_{i_k}, \mbf{S}_{j_k}$ are the Jacobians of $\mbf{c}_{ij}$ with respect to $\mc{X}_{i_k}, \mc{X}_{j_k}$, respectively. 

    \begin{itemize}
        \item On the reception of an input measurement $\mbf{u}_{i_{k-1}}$:
        \begin{align}
        \mccheck{X}_{i_k} &= \mbf{f}(\mchat{X}_{i_{k-1}}, \mbf{u}_{i_{k-1}}, \mbf{0}), \\
        \mbfcheck{P}_{i_k} &= \mbf{F}_{i_{k-1}} \mbfhat{P}_{i_{k-1}} \mbf{F}_{i_{k-1}}^\trans + \mbf{L}_{i_{k-1}}  \mbf{Q}_{k-1}  \mbf{L}_{i_{k-1}}^\trans.
        \end{align}
        \item On the reception of a local measurement $\mbf{y}_{i_k}$:
        \begin{align*} 
            \mbf{K} &= \mbfhat{P}_{i_{k}} \mbf{G}_{i_k}^\trans(\mbf{G}_{i_k} \mbfhat{P}_{i_{k}} \mbf{G}_{i_k}^\trans  + \mbf{R}_{i_k})^{-1},\\
            \delta \mbftilde{x} &= \mbf{K}(\mbf{y}_{i_k} - \mbf{g}(\mccheck{X}_{i_k})),\\
            \mctilde{X}_{i_k} &= \mccheck{X}_{i_k} \oplus \delta \mbftilde{x},\\
            \mbftilde{P}_{i_k} & = (\mbf{1} - \mbf{K} \mbf{G}_{i_k}) \mbfcheck{P}_{i_{k}}.
        \end{align*} 
        \item On the reception of a neighbors' state estimate $\tilde{\mc{X}}_{j_k}, \mbftilde{P}_{j_k}$:
        \begin{align*} 
            \mbftilde{P}_{i_k} &\gets \frac{1}{w} \mbftilde{P}_{i_k}, \qquad \mbftilde{P}_{j_k} \gets \frac{1}{1-w} \mbftilde{P}_{j_k}, \\
            \mbf{K} &= \mbftilde{P}_{i_k} \mbf{S}_{i_k}^\trans ( \mbs{\Psi} + \mbf{S}_{i_k} \mbftilde{P}_{i_k} ^\trans \mbf{S}_{i_k}^\trans+ \mbf{S}_{j_k} \mbftilde{P}_{j_k} ^\trans \mbf{S}_{j_k}^\trans)^{-1}, \\
            \delta \mbfhat{x} &= - \mbf{K}(\mbf{c}_{ij}(\mctilde{X}_{i_k}, \mctilde{X}_{j_k})), \\
            \mchat{X}_{i_k} &= \mctilde{X}_{i_k} \oplus \delta \mbfhat{x},\\
            \mbfhat{P}_{i_k} & = (\mbf{1} - \mbf{K} \mbf{G}_{i_k}) \mbftilde{P}_{i_{k}},
        \end{align*}
        and optionally further iterate \emph{both} robot estimates with \eqref{eq:general_mean}, \eqref{eq:general_cov}.

        \item At any time, send the current state estimate $\tilde{\mc{X}}_{j_k}, \mbftilde{P}_{j_k}$ to neighbors.
    \end{itemize}
\end{algorithm}

\diff{A feature of Algorithm \ref{alg:alg1} is that the robots can share their state information at anytime, with performance improving the more often sharing occurs, at the cost of increased communication bandwidth. After states are shared and pseudomeasurements are fused, the common states between robots will naturally drift due to sensor noise, until the next pseudomeasurement re-synchronizes the common states. This therefore becomes a tunable trade-off between estimation accuracy and communication bandwidth, which needs to be evaluated for a specific problem. In this paper's experiments, sharing is done at a set frequency of 10 Hz. Regarding computational complexity, Algorithm \ref{alg:alg1} will share a nearly-identical runtime to a Kalman filter.} 

\subsection{An Observability Test}
In a decentralized state estimation context, observability refers to the ability for \emph{each} robot to uniquely determine their state trajectory, given the inputs and measurements obtained by \emph{all} robots. \diff{Determining observability for a decentralized estimator is non-trivial due to the need to capture the communication topology within the test itself. To illustrate this, consider again the linear toy problem with two robots from Section \ref{sec:toy}. A naive approach to determining observability would be to construct one ``total state'' $\mbf{x} = [\mbf{x}_1 \;\; \mbf{x}_2]^\trans$, and to perform a standard observability test on this augmented system using the collected sensor measurements from both robots $y_1, y_2$. However, such a test would falsely conclude that the system is unobservable whereas the results from Section \ref{sec:toy} clearly show successful estimation. The pseudomeasurement must additionally be incorporated into the test to give the correct result, as the robots are reliant on communication to attain observabiliy of their individual states. To the best of the authors' knowledge, existing observability tests do not take into the full problem scope that this paper is concerned with. A decentralized observability test is presented in \cite{Pilloni2013} for linear-time-invariant systems with unknown input, but assumes each node is observable. The observability analysis in \cite{Huang2011a} has a similar approach to this paper, but is specific to their system where robots have identical state definitions.}

\diff{An advantage of the proposed approach is that the effects of communication of observability can be accurately captured by incorporating the pseudomeasurements themselves into a standard observability test. } \diff{Concretely, for nonlinear systems,} a \emph{local} observability test can be formed by considering the MAP problem on an entire trajectory simultaneously, but without prior information on the initial state (\cite{Psiaki2013}). \diff{Applying this to the multi-robot system,} let the bolded $\mbc{X}_k = (\mc{X}_{1_k}, \ldots, \mc{X}_{N_k})$ denote the ``total state'' of all robots at time step $k$, and $\mbf{y}_{k} = [\mbf{y}_{1_k}^\trans\; \ldots \; \mbf{y}_{N_k}^\trans]^\trans$ denote a stacked vector containing all the robots' local measurements at time step $k$. Let $\mbs{\psi}_k = [\;\ldots\; \mbf{y}_{ij}^\trans\; \ldots\;]^\trans, \; (i,j) \in \mc{E}$ denote the stacked pseudomeasurements between all robots at time step $k$. The MAP problem is 
\beq
\hat{\mbc{X}}_{0:K} = \argmax_{\mbc{X}_{0:K}} p(\mbc{X}_{0:K}| \mbf{y}_{0:K}, \mbs{\psi}_{0:K})
\eeq
with
\begin{multline}
    p(\mbc{X}_{0:K}| \mbf{y}_{0:K}, \mbs{\psi}_{0:K})= \\
    \eta \prod_{k=0}^K p(\mbf{y}_0|\mbc{X}_0) p(\mbs{\psi}_0|\mbc{X}_0) \prod_{k=1}^{K} p(\mbc{X}_k|\mbc{X}_{k-1}),
\end{multline}
leading to a nonlinear least squares problem with weight $\mbf{W}$ and error vector 
\begin{align*}
\mbf{e}(\mbc{X}_{0:K}) &= [\; \ldots \; \mbf{e}_{u,k}\; \ldots \;\mbf{e}_{y,k}\; \ldots \; \mbf{e}_{\psi,k}\; \ldots]^\trans,\\
\mbf{e}_{u,k} &= [\ldots (\mc{X}_{i_k} \ominus \mbf{f}(\mc{X}_{i_{k-1}}, \mbf{u}_{i_{k-1}}))^\trans \ldots], \\
\mbf{e}_{y,k} &= [\; \ldots \; (\mbf{y}_{i_k} - \mbf{g}(\mc{X}_{i_k}))^\trans \; \ldots \; ],  \quad i = 1, \ldots, N, \\
\mbf{e}_{\psi,k} & = [\; \ldots \; - \mbf{c}_{ij}(\mc{X}_{i_k}, \mc{X}_{j_k})^\trans \; \ldots \; ], \quad (i,j) \in \mc{E}.
\end{align*}
The error Jacobian is 
\begin{align} 
    \mbf{H} &=  \bma{cccc} -\mbf{F}_0 & \mbf{1} && \\ &\ddots & \ddots & \\ && -\mbf{F}_{K-1} & \mbf{1} \\ -\mbf{G}_0 &&& \\ & -\mbf{G}_1 && \\ && \ddots & \\ &&& -\mbf{G}_K\\ -\mbs{\Phi}_0 &&& \\ & -\mbs{\Phi}_1 && \\ && \ddots & \\ &&& -\mbs{\Phi}_K\ema, 
\end{align}
\begin{align}
    \mbf{F}_k &= \mathrm{diag}\left(\ldots, \frac{D \mbf{f}(\mc{X}_{i_k}, \mbf{u}_{i_k})}{D \mc{X}_{i_{k}}}, \ldots\right),\\
    \mbf{G}_k &= \mathrm{diag}\left(\ldots, \frac{D\mbf{g}(\mc{X}_{i_k})}{D \mc{X}_{i_{k}}}, \ldots \right), \quad i=1,\ldots, N,\\
    \mbs{\Phi}_{k} &= -\frac{D \mbf{e}_{\psi,k}(\mbc{X}_{k})}{D \mbc{X}_{k}},
\end{align}
with all undisplayed entries in $\mbf{H}$ equal to zero. For the solution to the MAP problem to be unique, then to first order, $(\mbf{H}^\trans \mbf{W} \mbf{H})$ must be invertible, and thus full rank. Fortunately, $\mbf{W}$ is always positive definite regardless of any cross-correlations that would add off-diagonal entries. Hence, 
$$\mathrm{rank}(\mbf{H}^\trans \mbf{W} \mbf{H}) = \mathrm{rank}(\mbf{H}),$$
and it is thus required that $\mbf{H}$ be full column rank. This implies that the proposed observability test is unaffected by the approximation induced by CI, or any cross-correlation terms that may or may not be successfully tracked. In a similar way to \cite[Ch~3.1.4]{Barfoot2022}, it can be shown that by performing a variety of elementary row/column operations, the rank of $\mbf{H}$ is equivalent to the rank of 
\beq 
\mc{O} = \bma{c} \mbf{M}_0 \\ \mbf{M}_1 \mbf{F}_0 \\ \vdots \\ \mbf{M}_K \mbf{F}_{K-1} \ldots \mbf{F}_0 \ema, \qquad \mbf{M}_k = \bma{c} \mbf{G}_k \\ \mbs{\Phi}_k \ema.
\eeq
Hence, if $\mc{O}$ has maximum rank, the solution to the MAP problem is locally unique, and the system is said to be observable. Note that this test easily allows for time-varying graphs, which would yield a different $\mbs{\Phi}_k$ for each time step $k$.

\section{Efficient Odometry Sharing using Preintegration} \label{sec:preintegration}
Many problems, especially those where robots estimate their neighbors' positions, will require robots to have access to their neighbors' process-model input values $\mbf{u}$. \diff{Previously in this paper}, it has been assumed that all robots have unrestricted access to each other's inputs. In robot state estimation applications, the input is often the odometry measurements, such as wheel encoder or IMU measurements. These can occur at frequencies of 100 - 1000~Hz, and can therefore be infeasible to share in real time, especially if multiple robots are to simultaneously share measurements at high frequency. This could quickly reach a bandwidth limit on the common communication channel, such as ultra-wideband radio.  \diff{While \cite{Xu2020} solve this problem by directly sharing pose changes between two points in time, this violates statistical independence assumptions and leads to inconsistent estimates.}  

The proposed solution to this problem is to use \emph{preintegration}. That is, robots will instead share preintegrated input measurements over an arbitrary duration of time instead of individual input measurements. Specifically, consider the following generic process model 
\beq 
\mc{X}_k = \mbf{f}(\mc{X}_{k-1}, \mbf{u}_{k-1}, \mbf{w}_{k-1}).
\eeq
 The action of preintegration is to directly iterate this process model by repeated compositions in order to, after algebraic manipulation, generate a new \emph{preintegrated process model} $\mbf{f}_{pq}$ that relates two states at arbitrary time steps $k=p$ and $k=q$. That is,
\begin{align}
    \mc{X}_{p+1} &= \mbf{f}(\mc{X}_p, \mbf{u}_p, \mbf{w}_p),\\
    \mc{X}_{p+2} &= \mbf{f}(\mbf{f}(\mc{X}_p, \mbf{u}_p, \mbf{w}_p), \mbf{u}_{p+1}, \mbf{w}_{p+1}), \\
    & \vdots \\
    \mc{X}_q &= \mbf{f}(\mbf{f}(\;\ldots\;\mbf{f}(\mc{X}_p, \mbf{u}_p, \mbf{w}_p)\;\ldots\;),\mbf{u}_{q-1}, \mbf{w}_{q-1}) 
    \\
    &\triangleq \mbf{f}_{pq}(\mc{X}_i, \Delta \mc{X}_{pq}) \oplus \mbf{w}_{pq} \label{eq:preint3},
\end{align}
where $\Delta \mc{X}_{pq}$ is the \emph{relative motion increment} (RMI), which in general may also belong to a Lie group, and $\mbf{w}_{pq}\sim \mc{N}(\mbf{0}, \mbf{Q}_{pq})$ is the \emph{preintegrated noise}. The advantages of preintegration will stem from the careful choice of RMI definition, which is ideally done such that the RMI has the following properties.
\begin{enumerate}
    \item The RMI is determined from the input measurements exclusively, and is independent of the state estimate:
    \beq 
    \Delta \mc{X}_{pq} = \Delta \mc{X}_{pq}(\mbf{u}_{p:q-1}).
    \eeq 
    \item Far fewer numbers are required to represent the RMI than the $(q-p)$ raw measurements that occurred during the preintegration interval:
    \beq 
    \dim(\Delta \mc{X}_{pq}) << (q-p)\dim(\mbf{u}_k).
    \eeq 
\end{enumerate}
If the above points are true, communicating $\Delta \mc{X}_{pq}, \mbf{Q}_{pq}$ instead of $\mbf{u}_{p:q-1}$ will not only reduce the communication cost, but will also result in a fixed message size and ability to directly predict the state forward over a long duration of time, instead of sequentially processing the measurements. \diff{Note that it is not always possible to define an RMI such that \eqref{eq:preint3} holds exactly. However,} it turns out that many common process models in robotics are amenable to preintegration (\cite{Barrau2019, Eckenhoff2019}), and furthermore are typically extremely fast to preintegrate incrementally as input measurements are obtained (\cite{Forster2017}). In other words, there exists a function $\textsc{Increment}(\cdot)$ \diff{defined such that it satisfies}
$$ 
\Delta \mc{X}_{pq}, \mbf{Q}_{pq} = \textsc{Increment}(\Delta \mc{X}_{p:q-1}, \mbf{Q}_{p:q-1}, \mbf{u}_{q-1}).
$$
A few examples now follow, which describe concrete implementations of $\Delta \mc{X}_{pq}$, $\mbf{f}_{pq}(\cdot)$, and $\textsc{Increment}(\cdot)$.


\begin{example}[Linear preintegration]
    The linear process model
    \beq 
    \mbf{x}_k = \mbf{F}_{k-1} \mbf{x}_{k-1} + \mbf{L}_{k-1}\mbf{u}_{k-1}
    \eeq
    can be directly iterated to yield
    \begin{align}
    \mbf{x}_q &= \left(\prod_{k=p}^{q-1} \mbf{F}_k\right) \mbf{x}_p + \sum_{k=p}^{q-1} \left(\prod_{\ell = k + 1}^{q-1} \mbf{F}_\ell\right) \mbf{L}_k \mbf{u}_k \\ 
    &\triangleq \mbf{F}_{pq} \mbf{x}_p + \Delta \mbf{x}_{pq}, \label{eq:linear_preint}
    \end{align}
    where \eqref{eq:linear_preint} defines $\mbf{f}(\cdot)$. Assuming that noise enters the model additively through the input $\mbf{u}_{k} = \mbfbar{u}_k + \mbf{w}_k$, where $\mbf{w}_k \sim \mc{N}(\mbf{0}, \mbf{Q}_k)$, \eqref{eq:linear_preint} becomes $\mbf{x}_j = \mbf{F}_{pq} \mbf{x}_i + \Delta \mbfbar{x}_{pq} + \mbf{w}_{pq}$ where
    \begin{align} 
        \mbf{w}_{pq}& \triangleq \sum_{k=p}^{q-1} \left(\prod_{\ell = k + 1}^{q-1} \mbf{F}_\ell\right) \mbf{L}_k \mbf{w}_k ,\\      
        &= \mbf{F}_{q-1} \mbf{w}_{pq-1} + \mbf{L}_{q-1} \mbf{w}_{q-1}.
    \end{align} 
    The RMI $\Delta \mbf{x}_{pq}$ and corresponding covariance are therefore built incrementally with
    \begin{align} 
        \Delta \mbf{x}_{pq} &= \mbf{F}_{q-1} \Delta \mbf{x}_{pq-1} + \mbf{L}_{q-1} \mbf{u}_{q-1},\\
    \mbf{Q}_{pq} &= \mbf{F}_{q-1} \mbf{Q}_{pq-1} \mbf{F}_{q-1}^\trans + \mbf{L}_{q-1} \mbf{Q}_{q-1} \mbf{L}_{q-1}^\trans,
    \end{align}
   which together define the $\textsc{Increment}(\cdot)$ function.
    

\end{example}

\begin{example}[Wheel odometry preintegration on $SE(2)$] \label{ex:wheel_odometry}
    
Given a robot pose $\mbf{T} \in SE(2)$, the wheel odometry process model is given by 
\beq 
\mbf{T}_k = \mbf{T}_{k-1} \Exp(\Delta t \mbf{u}_{k-1}),
\eeq 
where $\mbf{u} = [\omega\; v\;0]^\trans$, $\omega$ is the robot's heading rate-of-change, and $v$ is its forward velocity in its own body frame. Direct iteration yields the preintegrated process model $\mbf{f}_{pq}(\cdot)$ given by 
\beq 
\mbf{T}_q = \mbf{T}_p \underbrace{\prod_{k=p}^{q-1} \Exp(\Delta t\mbf{u}_k)}_{\triangleq \Delta \mbf{T}_{pq}}.
\eeq
Noise $\mbf{w}_k \sim \mc{N}(\mbf{0}, \mbf{Q}_k)$ is again assummed to enter additively through the input, and a series of first-order approximations lead to
\begin{align*}
    \Delta \mbf{T}_{pq} &= \prod_{k=p}^{q-1} \Exp(\Delta t(\mbfbar{u}_k + \mbf{w}_k)) \\
    &\approx \prod_{k=p}^{q-1} \Exp(\Delta t\mbfbar{u}_k) \Exp(\Delta t\mbf{J}_k \mbf{w}_k) \\
    & \approx  \Delta \mbfbar{T}_{pq} \underbrace{\prod_{k=p}^{q-1} \Exp(\Delta t\mathbf{Ad}(\Delta \mbf{T}_{k+1j}^{-1})\mbf{J}_k \mbf{w}_k)}_{\Exp(\mbf{w}_{pq})},
\end{align*}
where $\mbf{J}_k \triangleq \mbf{J}(\Delta t\mbfbar{u}_k)$ is the right Jacobian of $SE(2)$. Having identified an expression for $\Exp(\mbf{w}_{pq})$, under the assumption that $\mbf{w}_{pq}$ is small, 
\begin{align*} 
\mbf{w}_{pq} &\approx \sum_{k=p}^{q-1} \Delta t\mathbf{Ad}(\Delta \mbf{T}_{k+1q}^{-1}) \mbf{J}_k \mbf{w}_k \\
& = \sum_{k=p}^{q-2}\Delta t \mathbf{Ad}(\Delta \mbf{T}_{k+1q}^{-1}) \mbf{J}_k \mbf{w}_k \\
& \hspace{2cm}+ \Delta t\underbrace{\mathbf{Ad}(\Delta \mbf{T}_{qq}^{-1})}_{\mbf{1}} \mbf{J}_{q-1} \mbf{w}_{q-1} \\
&= \underbrace{\mathbf{Ad}(\Delta \mbf{T}_{q-1q}^{-1})}_{\triangleq \mbf{F}_{q-1}} \mbf{w}_{pq-1}+  \underbrace{\Delta t\mbf{J}_{q-1}}_{\triangleq \mbf{L}_{q-1}} \mbf{w}_{q-1},
\end{align*}
and the defining operations of the $\textsc{Increment}(\cdot)$ function follow,
\begin{align} 
    \Delta \mbf{T}_{pq } &= \Delta \mbf{T}_{pq-1} \Exp(\Delta t \mbf{u}_k),  \\
\mbf{Q}_{pq} &= \mbf{F}_{q-1} \mbf{Q}_{pq-1} \mbf{F}_{q-1}^\trans + \mbf{L}_{q-1} \mbf{Q}_{q-1} \mbf{L}_{q-1}^\trans.
\end{align}
\end{example}

\begin{example}[IMU preintegration] Being the most well-known usage of preintegration, a complete reference for IMU preintegration on the $SO(3) \times \mathbb{R}^3 \times \mathbb{R}^3$ manifold can be obtained from \cite{Forster2017}, and alternatively for the $SE_2(3)$ group from \cite{Brossard2020b, Barfoot2022}. Either approach can be used with the framework in this paper. However in Section \ref{sec:quads} of this paper, $\mbf{T}_{wi}  \in SE_2(3)$ matrices are used to represent the extended pose of Robot $i$ relative to an inertial world frame $w$. Following \cite{Shalaby2023} and \cite{Brossard2020b}, it can be shown that the discrete-time IMU kinematic equations can be written in the form 
    \beq 
    \mbf{T}_{wi_k} = \mbf{G}_{k-1} \mbf{T}_{wi_{k-1}} \mbf{U}_{k-1}, \label{eq:imu_process}
    \eeq 
where
\begin{align*}
\mbf{T}_{wi_k} &= \bma{ccc} \mbf{C} & \mbf{v} & \mbf{r} \\ \mbf{0} & 1 & 0 \\ \mbf{0} & 0 & 1 \ema \\
\mbf{G}_{k-1} &= \bma{ccc} \mbf{1} & \Delta t \mbf{g} & -\frac{\Delta t^2}{2}  \mbf{g} \\ 0 & 1 & - \Delta t \\ 0 & 0 & 1 \ema \\
\mbf{U}_{k-1} &= \bma{ccc} \exp(\Delta t \mbs{\omega}^{\wedge}) & \Delta t\mbf{J}(\Delta t \mbs{\omega}) \mbf{a} & \frac{\Delta t^2}{2} \mbf{N}(\Delta t \mbs{\omega}) \mbf{a} \\ 0 & 1 & \Delta t \\ 0 & 0 & 1 \ema \\ 
\mbf{N}(\mbs{\phi}) &=\mbf{z} \mbf{z}^\trans + 2\left(\frac{1}{\phi} - \frac{\sin \phi}{\phi^2}\right) \mbf{z}^\wedge + 2\frac{\cos \phi - 1}{\phi^2} \mbf{z}^\wedge \mbf{z}^\wedge \\
\phi& = \norm{\mbs{\phi}}, \; \mbf{z} = \mbs{\phi}/\phi,
\end{align*}
and $\mbf{C} \in SO(3),\; \mbf{v},\; \mbf{r}$ respectively represent attitude, velocity, position relative to the world frame $w$, $\mbs{\omega}$ is the IMU's unbiased gyro measurement, $\mbf{a}$ is the IMU's unbiased accelerometer measurement, $\mbf{g}$ is the gravity vector resolved in frame $w$, and $\mbf{J}(\mbs{\phi})$ is the left Jacobian of $SO(3)$. Preintegration of these kinematics is easily achieved by direct iteration with
\begin{align} 
\mbf{T}_{wi_{q}} &= \left(\prod_{k=p}^{q-1} \mbf{G}_{k-1}\right) \mbf{T}_{wi_p} \left(\prod_{k=p}^{q-1} \mbf{U}_{k-1}\right) \\
&\triangleq \Delta \mbf{G}_{pq} \mbf{T}_{wi_p} \Delta \mbf{U}_{pq},
\end{align}
where $\Delta \mbf{U}_{pq}$ is the RMI, and the noise statistics can be propagated through this preintegration process by a standard linearization procedure. \cite{Shalaby2023} present the details of the noise propagation, as well as how to adapt this formulation for relative poses.
\end{example}

\subsection{Multi-robot preintegration}

In the context of multi-robot estimation problems, an individual robot's process model may involve the input values of many neighboring robots. To reflect this, rewrite the process model for Robot $i$ as 
\beq 
\mc{X}_{i_k} = \mbf{f}(\mc{X}_{i_{k-1}}, \mbf{u}_{i_{k-1}}, \mbf{u}_{j_{k-1}}), \qquad j \in \mc{N}_i, \label{eq:multi_input1}
\eeq 
where $\mbf{u}_{i_{k}}$ denotes an input measured by Robot $i$ and $\mc{N}_i$ denotes the set of neighbor IDs of Robot $i$. The preintegrated process model would now be written as 
\beq 
\mc{X}_{i_q} = \mbf{f}_{pq}(\mc{X}_{i_p}, \Delta \mc{X}_{i_{pq}}, \Delta \mc{X}_{j_{pq}}), \qquad j \in \mc{N}_i,\label{eq:multi_input2}
\eeq
where $ \Delta \mc{X}_{i_{pq}}$ denotes an RMI calculated from the input measurements of Robot $i$. 

A complication is that the RMIs from neighboring Robots $\Delta \mc{X}_{j_{pq}}$ are only available asynchronously, meaning it is not always possible to evaluate \eqref{eq:multi_input2} directly. To deal with this, assume that \diff{the state can be temporarily partially propagated with a value of $\mbf{0}$ for the neighbor's input, and then fully propagated once the RMI is shared.} That is, assume the preintegrated process model $\mbf{f}_{pq}$ is compatible with
\begin{align} 
    \mathfrak{X}_{i_{k-1}} &= \mbf{f}(\mc{X}_{i_{k-2}}, \mbf{u}_{i_{k-2}}, \mbf{0}), \\
    \mathfrak{X}_{i_k} &= \mbf{f}(\mathfrak{X}_{i_{k-1}}, \mbf{u}_{i_{k-1}}, \mbf{0}), \\
    \mc{X}_{i_k} &= \mbf{f}_{pq}(\mathfrak{X}_{i_k}, \mc{I}, \Delta \mc{X}_{j_{pq}} ),
\end{align} 
where $\mc{I}$ is the ``identity'' or ``zero'' RMI \diff{constructed from input values of zero, and $\mbf{f}$ is defined as per \eqref{eq:multi_input1} with $\mbf{u}_{j}$ substituted for $\mbf{0}$}. The variable $\mathfrak{X}_{i_k}$ represents an intermediate, non-physical state that is propagated using the process model without input information from neighboring robots. Sometime later, at arbitrary time step $k=q$, the RMI from a neighboring robot $\Delta \mc{X}_{j_{pq}}$ is received and this intermediate state $\mathfrak{X}_{i_{q}}$ is propagated back into a physically meaningful quantity $\mc{X}_{i_{q}}$. A concrete example of this asynchronous intermediate state updating is shown in Section \ref{sec:ground_robots}, and a summary is shown in Algorithm \ref{alg:alg2}.

\begin{algorithm}[H]
    \caption{Decentralized estimation with preintegration.}\label{alg:alg2}
    The setup is identical to Algorithm \ref{alg:alg1}, except that the process model requires input information from neighboring robots as in \eqref{eq:multi_input1}, \eqref{eq:multi_input2}.
    \begin{itemize}
        \item On the reception of a local input measurement $\mbf{u}_{i_{k-1}}$:
        \begin{align*} 
        {\mathfrak{X}}_{i_k} &= \mbf{f}(\mchat{X}_{i_{k-1}}, \mbf{u}_{i_{k-1}}, \mbf{0}) \\
        \mathfrak{P}_{i_k} &= \mbf{F}_{i_{k-1}} \mbfhat{P}_{i_{k-1}} \mbf{F}_{i_{k-1}}^\trans + \mbf{L}_{i_{k-1}}  \mbf{Q}_{k-1}  \mbf{L}_{i_{k-1}}^\trans., 
        \end{align*}
        and increment Robot $i$'s own RMI,
        \bdis 
        \Delta \mc{X}_{i_{pk}}, \mbf{Q}_{i_{pk}} = \textsc{Increment}(\Delta \mc{X}_{i_{pk-1}}, \mbf{Q}_{i_{pk-1}}, \mbf{u}_{i_{k-1}}).
        \edis
        \item Whenever required, send  $\Delta \mc{X}_{i_{pq}}, \mbf{Q}_{i_{pq}}$ to neighbors.
        \item On the reception of a neighboring robot's RMI and covariance $\Delta \mc{X}_{j_{pq}}, \mbf{Q}_{j_{pq}}$,
        \begin{align*} 
        \mc{X}_{i_q} &= \mbf{f}_{pq}(\mathfrak{X}_{i_q}, \mc{I}, \Delta \mc{X}_{j_{pq}} ), \\
        \mbf{F}_{pq} &= \left.\frac{D \mbf{f}_{pq}(\mathfrak{X}, \mc{I}, \Delta \mc{X}_{j_{pq}} )}{D \mathfrak{X}}\right|_{\mathfrak{X}_{i_q}},\\
        \mbfcheck{P}_{i_q} &=  \mbf{F}_{pq} \mathfrak{P}_{i_{k-1}}  \mbf{F}_{pq}^\trans + \mbf{Q}_{j_{pq}}. 
        \end{align*}
        \item On the reception of a local measurement $\mbf{y}_{i_k}$, proceed as per Algorithm \ref{alg:alg1}.
        \item On the reception of a neighbors' state estimate $\tilde{\mc{X}}_j, \mbftilde{P}_j$, proceed as per Algorithm \ref{alg:alg1}.
        \item At any time, send the current state estimate $\tilde{\mc{X}}_{j_k}, \mbftilde{P}_{j_k}$ to neighbors.
    \end{itemize}
\end{algorithm}

\subsection{Estimating Input Biases} \label{sec:biases} For some problems, it may be desired to estimate an input bias $\mbf{b}$ as part of the overall state $\mc{X}$, a setup commonly occuring in inertial navigation where accelerometer and rate gyro biases are estimated. The difficulty lies in the frequent inability to express RMIs independently of the bias values, thus leaving RMIs in the form of 
\beq 
\Delta \mc{X}_{pq}(\mbf{u}_{p:q-1}, \mbf{b}_{p:q-1}).
\eeq
In the context of the multi-robot estimation scheme shown in Algorithm \ref{alg:alg2}, computing RMIs this way causes inconsistency in the filter, since the RMIs are now correlated with the robot states. Accounting for this would require maintaining  the cross-correlation between a robot's state and their neighbors' biases. 

A simpler alternate solution is to have robots estimate their neighbors' input biases in addition to their own. This requires to exploit the fact that biases are usually modelled to follow a random walk, and therefore have a constant mean in the absence of any correcting information. This motivates the approximation $\mbf{b}_p \approx \mbf{b}_{p+1} \approx \ldots \approx \mbf{b}_{q}$ and hence
\beq 
\Delta \mc{X}_{pq}(\mbf{u}_{p:q-1}, \mbf{b}_{p:q-1}) \approx 
\Delta \mc{X}_{pq}(\mbf{u}_{p:q-1}, \mbf{b}_{q}).
\eeq
When robots receive input measurements, they increment their RMIs with raw (biased) inputs to produce \diff{biased RMIs} $\Delta \mc{X}_{pq}(\mbf{u}_{p:q-1}, \mbf{0})$. At an appropriate time, they share their current biased RMIs, which is corrected for bias by the receiving robot using the first-order approximation
\begin{align}
\Delta \mc{X}_{pq}(\mbf{u}_{p:q-1}, \mbf{b}_{q}) &\approx \Delta \mc{X}_{pq}(\mbf{u}_{p:q-1}, \mbf{0}) \oplus \mbf{B}_{pq} \mbf{b}_q, \label{eq:rmi_bias_correction}\\
\mbf{B}_{pq} &\triangleq \frac{D}{D \mbf{b}_q} \Big(\Delta \mc{X}_{pq}(\mbf{u}_{p:q-1}, \mbf{b}_{q})\Big) ,
\end{align}
where $\mbf{B}_{pq}$ is defined as the bias Jacobian. Equation \eqref{eq:rmi_bias_correction} is an approximation that contributes unmodelled errors to the estimation problem, relying on an assumption that $\mbf{b}_q$ is small. This can be enabled by a proper offline calibration procedure that removes any large bias, and only small deviations from this are estimated online. 
Such a procedure is used for the quadcopter problem in Section \ref{sec:quads}, where good, consistent estimation results are still obtained despite the approximation in \eqref{eq:rmi_bias_correction}.


\subsection{Autoencoding Covariance Matrices}
\label{sec:preint_autoencoding}
As shown in Algorithm \ref{alg:alg2}, predicting state estimates that are a function of neighbor inputs requires the RMI $\Delta \mc{X}_{pq}$ along with a corresponding covariance $\mbf{Q}_{pq}$. As is, these two quantities must be shared between robots. This section proposes an optional method that further reduces communication costs by eliminating the requirement to share the preintegrated  covariance $\mbf{Q}_{pq}$.

\begin{figure}[b]
    \centering 
    \includegraphics[width=\linewidth, clip, trim={0 6cm 0.0cm, 2cm}]{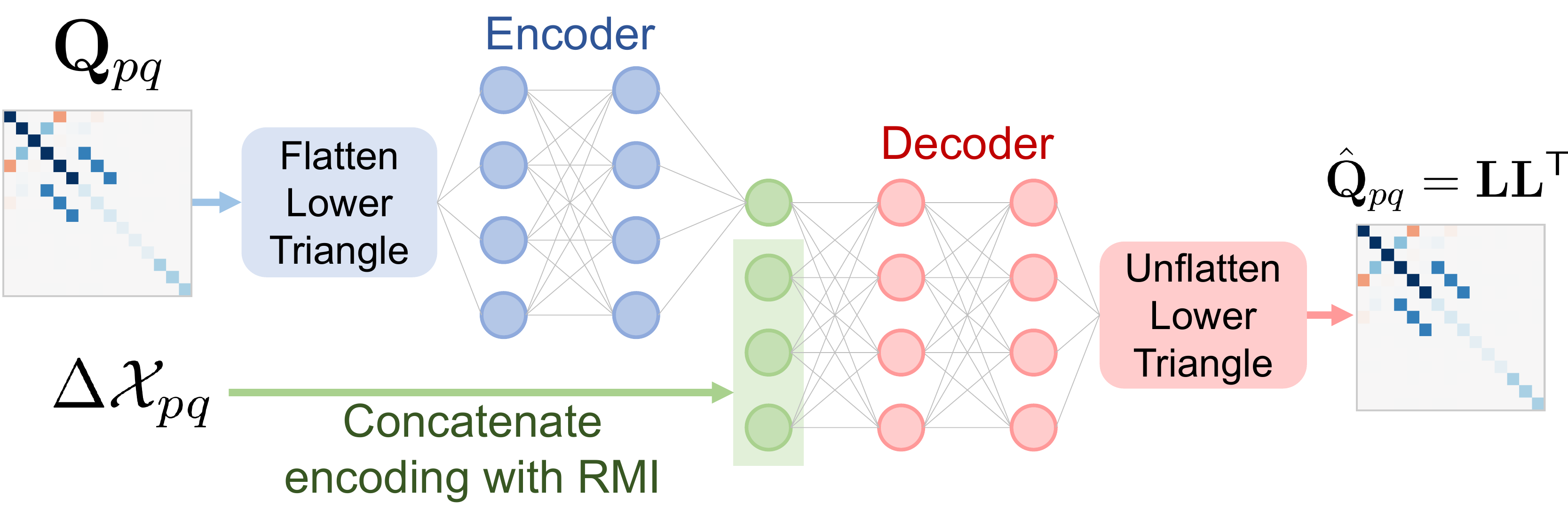}
    \caption{\diff{Concept diagram of mean-assisted autoencoder.}}
    \label{fig:mae_diagram}
\end{figure}

The key insight is that $\Delta \mc{X}_{pq}(\mbf{u}_{p:q-1})$ and $\mbf{Q}_{pq}(\mbf{u}_{p:q-1})$ are both calculated from the same input values $\mbf{u}_{p:q-1}$. Hence, if an alternate mapping $\mbf{Q}_{pq} = \mbf{h}(\Delta \mc{X}_{pq})$ existed, then it would be sufficient to share $\Delta \mc{X}_{pq}$ only, and the receiving robot could infer $\mbf{Q}_{pq}$ directly from the RMI. In the absence of analytic expressions for $\mbf{h}(\cdot)$, this paper approximates the function with a neural network, trained on purely synthetic RMI-covariance pairs. An additional complication is that such a function $\mbf{h}(\cdot)$ may not always exist since an RMI can correspond to many possible covariances depending on the input values. In this case $\mbf{h}(\cdot)$ is not a true function since it is one-to-many, and its definition is modified to also accept a low-dimensional \emph{encoding} $\mbs{e}(\mbf{Q}_{pq})$, leading to $\mbf{Q}_{pq} = \mbf{h}(\Delta \mc{X}_{pq}, \mbs{e}(\mbf{Q}_{pq}))$. This leads to an architecture here referred to as \emph{mean-assisted autoencoding}, depicted in Figure \ref{fig:mae_diagram}.
\begin{figure}
    \centering 
    \includegraphics[width=\linewidth]{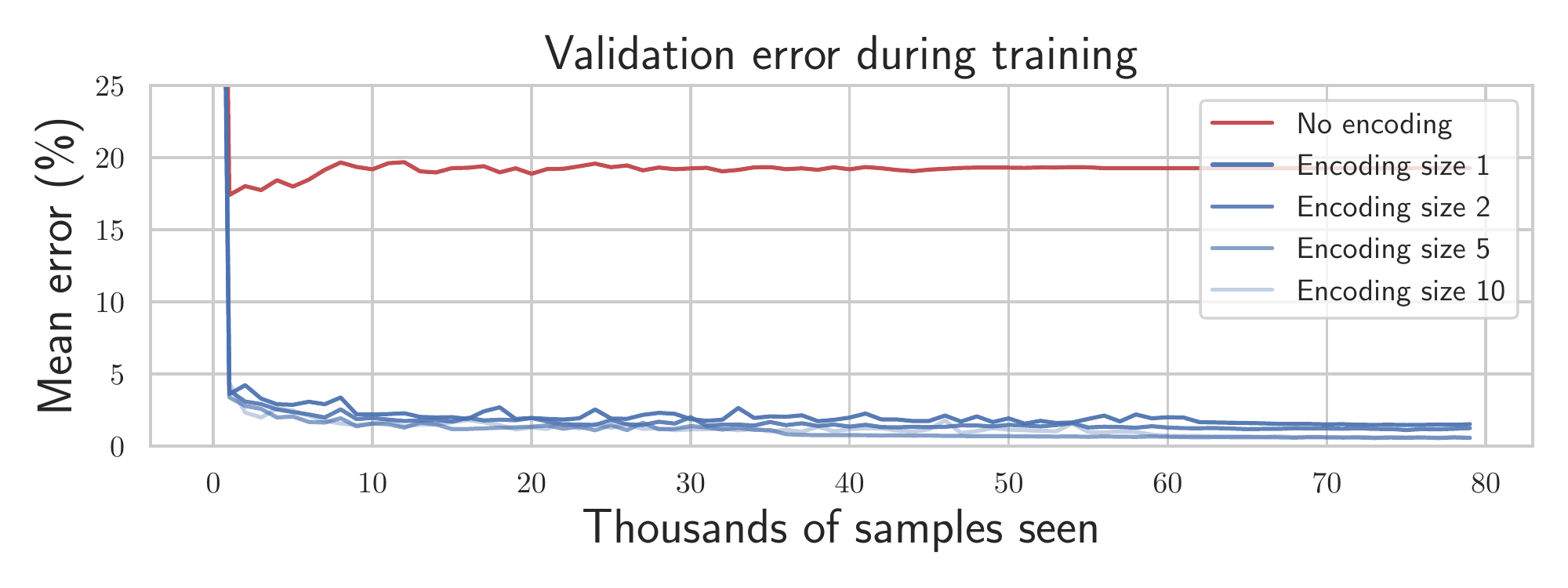}
    \caption{Mean percentage reconstruction error throughout training for various encoding sizes including no encoding. A single encoding number is sufficient to achieve less than 1\% reconstruction error on average.}
    \label{fig:rmi_convergence}
\end{figure}
\begin{figure}
    \centering 
    \includegraphics[width=\linewidth]{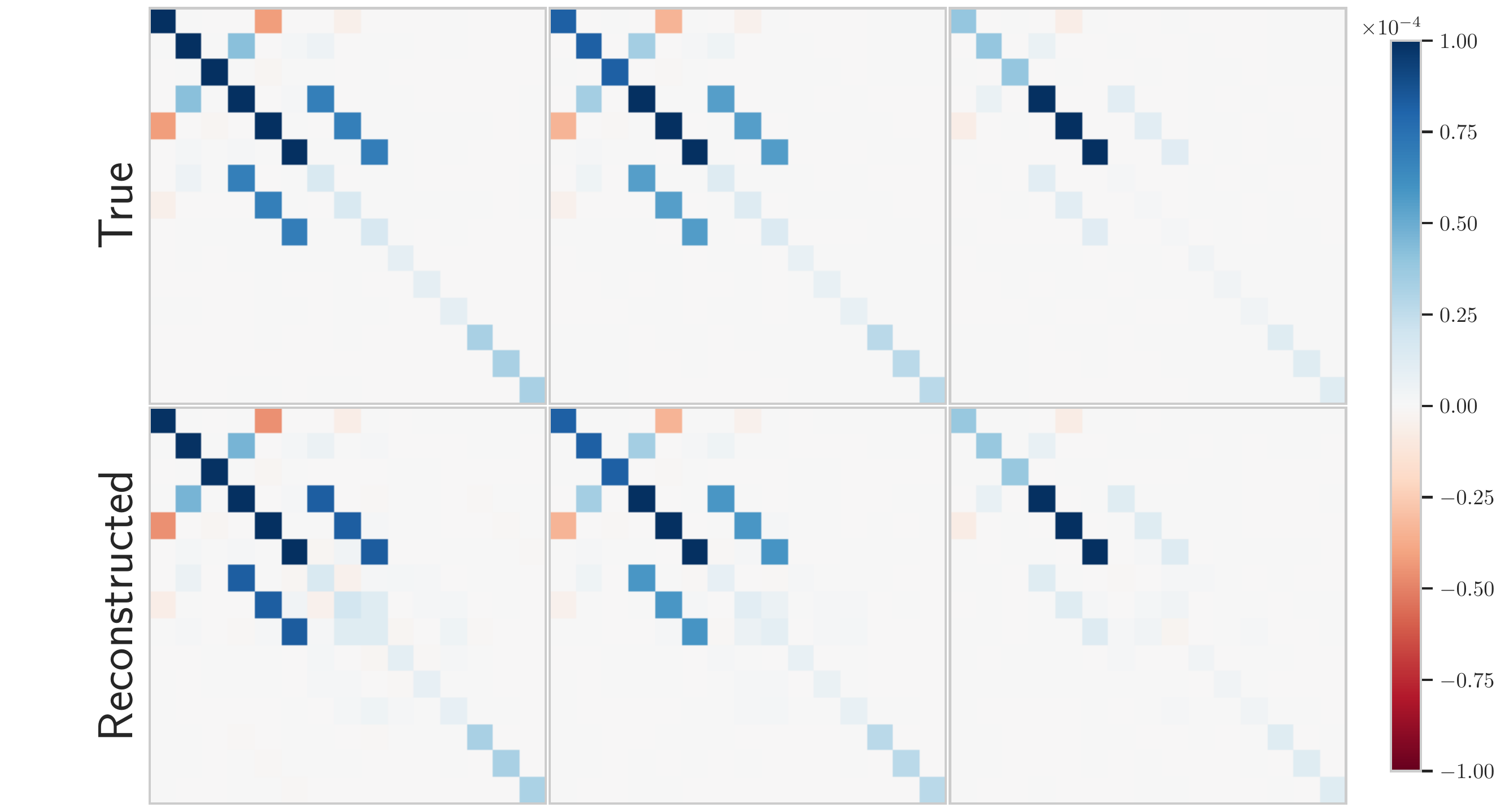}
    \caption{Visualization of preintegrated IMU noise covariance matrices along with reconstruction using mean-assisted autoencoding.}
    \label{fig:rmi_reconstruction}
\end{figure}
\begin{figure*}[t]
    \centering
    \includegraphics[width=0.5\linewidth]{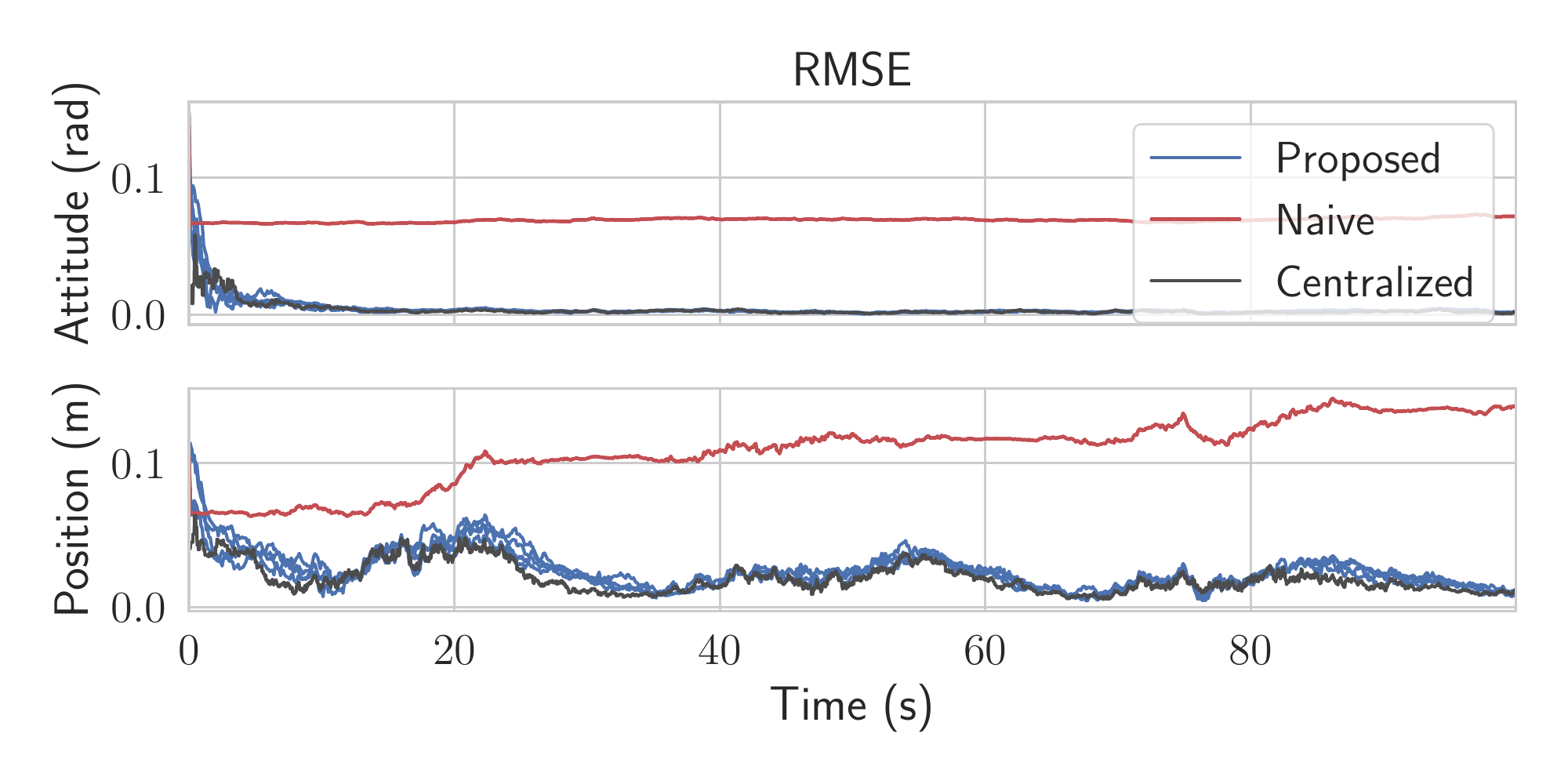}%
    \includegraphics[width=0.5\linewidth]{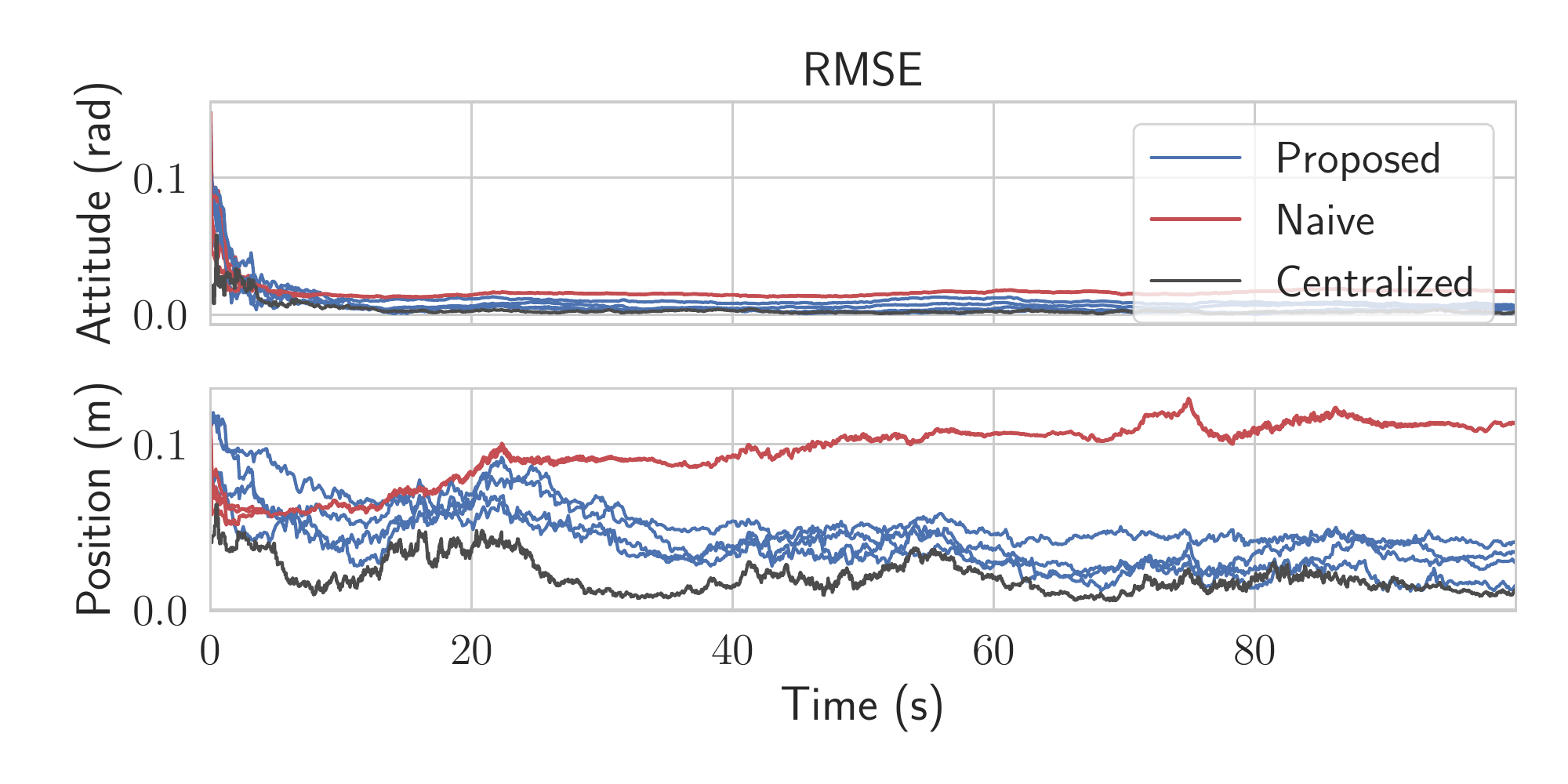}
    \caption{RMSE for the ground robot simulation. There are four blue lines for the four robots running the proposed algorithm, and four visibly coincident red lines for the naive algorithm. \textbf{Left:} State fusion occuring at 10~Hz. \textbf{Right:} State fusion occuring at 1~Hz.}
    \label{fig:se2_rmse}
\end{figure*}   
\begin{figure}[t]
    \centering
    \includegraphics[width=\linewidth]{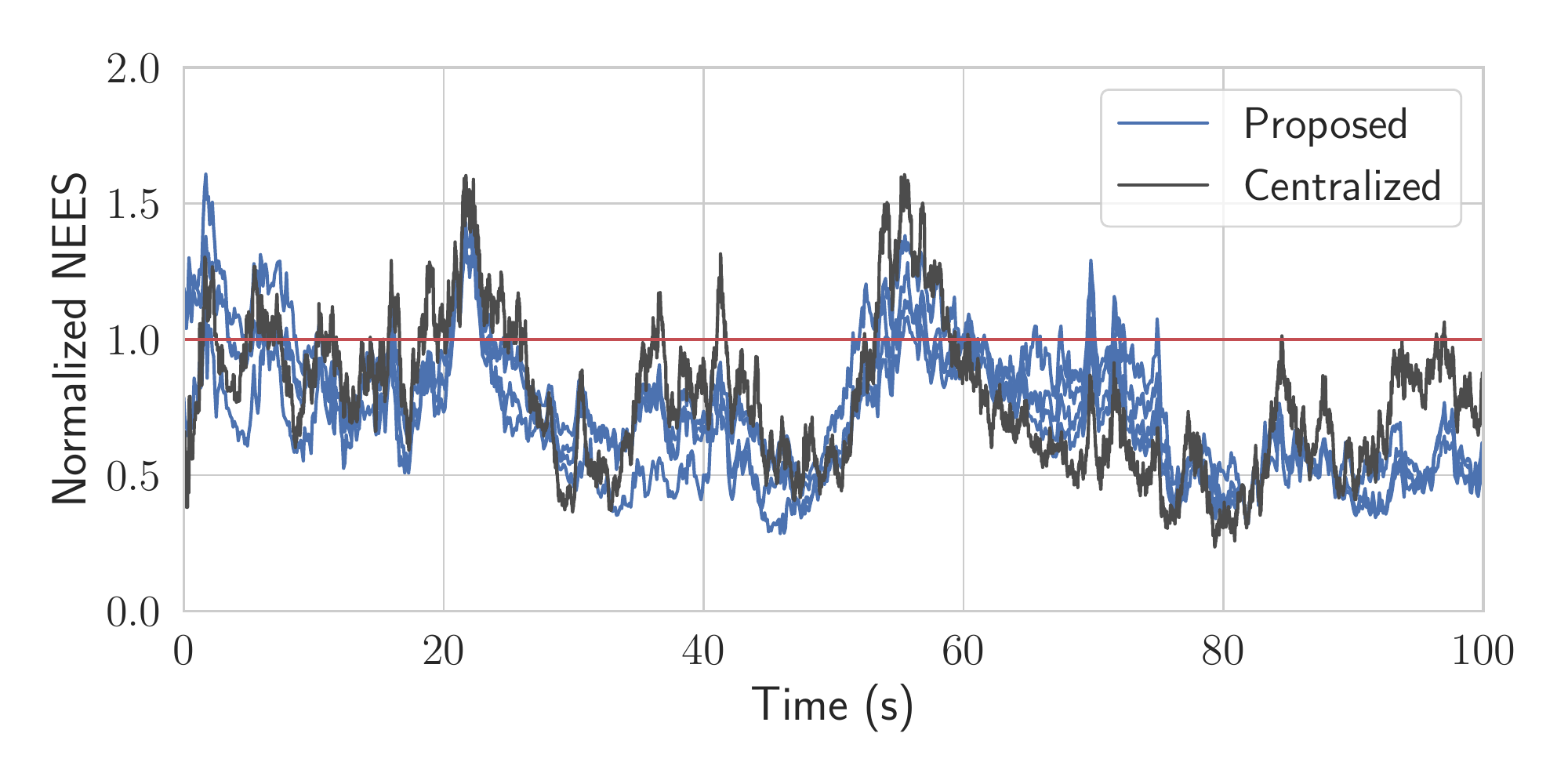}
    \caption{\diff{50-trial NEES plot for the ground robot simulation for the proposed vs. centralized solution, with the multiple blue lines each representing a robot. The naive solution without CI is far outside the plot. The red line represents the expected NEES value.}}
    \label{fig:se2_nees}
\end{figure}

The flattened lower-triangular half of $\mbf{Q}_{pq}$ is given to a simple fully-connected \emph{encoder} network with GELU activation functions and a single hidden layer with 256 neurons. The output of this network is the encoding, which can be as small as one or two numbers. This encoding is then concatenated with a parameterization of the RMI $\Delta \mc{X}_{pq}$ and fed to a similar \emph{decoder} network, again with a single 256-neuron hidden layer. \diff{The decoder network outputs the flattened lower-triangular half of a Cholesky decomposition $\mbf{L}$, which is used to reconstruct the matrix using $\mbfhat{Q}_{pq} = \mbf{L}\mbf{L}^\trans$. The covariance matrix is guaranteed to be positive definite as long as the diagonal elements of $\mbf{L}$ are non-zero, which is extremely unlikely to occur in practice.} For training, the loss function simply uses the Frobenius norm,
\beq 
\mc{L}(\mbf{Q}_{pq}, \mbfhat{Q}_{pq}) = ||\mbf{Q}_{pq} - \mbfhat{Q}_{pq}||_{\frob}.
\eeq
Figure \ref{fig:rmi_convergence} shows the training convergence history for various encoding sizes, applied to IMU preintegration. The Adam optimizer is chosen with an initial learning rate of $10^{-3}$ that is scheduled to decrease once the loss plateaus. The training process takes just a few minutes on a laptop CPU to achieve less than 1\% average reconstruction error on a validation dataset from experimental data. \diff{The ``No encoding'' baseline represents an architecture where the decoder network attempts to infer the covariance matrix directly from the RMI itself, without using the output of the encoder network, and hence eliminating the need to communicate RMI covariance information. However, doing this yields substantially higher error than when an encoding is provided.}  A visualization of the reconstructed covariance matrices can be seen in Figure \ref{fig:rmi_reconstruction}, using an encoding size of only 1 number. Section \ref{sec:quads} will employ this method ``in the loop''  for a real quadcopter problem. It will be shown that the reconstruction error is so small that the impact on the estimation results are negligible. 

The training data is purely synthetic, where RMIs are constructed from a random amount of random IMU measurements, with values covering the realistic range of real IMU measurements. Since the length of the dataset is infinite, the risk of overfitting is completely eliminated, \diff{as long as the real IMU measurements lie within the range of randomly generated values. In fact, the networks immediately generalize to any physical sensor of the same type. Physical characteristics such as biases, scale factors, and axis-misalignments are irrelevant since the result after these effects is still a list of values representing the sensor measurements. As long as those values remain within the randomly-generated training domain, the network will perform well. In Section \ref{sec:quads}, the same autoencoder is used on three different quadcopters each with a different physical IMUs.}

\begin{figure}
    \centering
    \includegraphics[width=\linewidth]{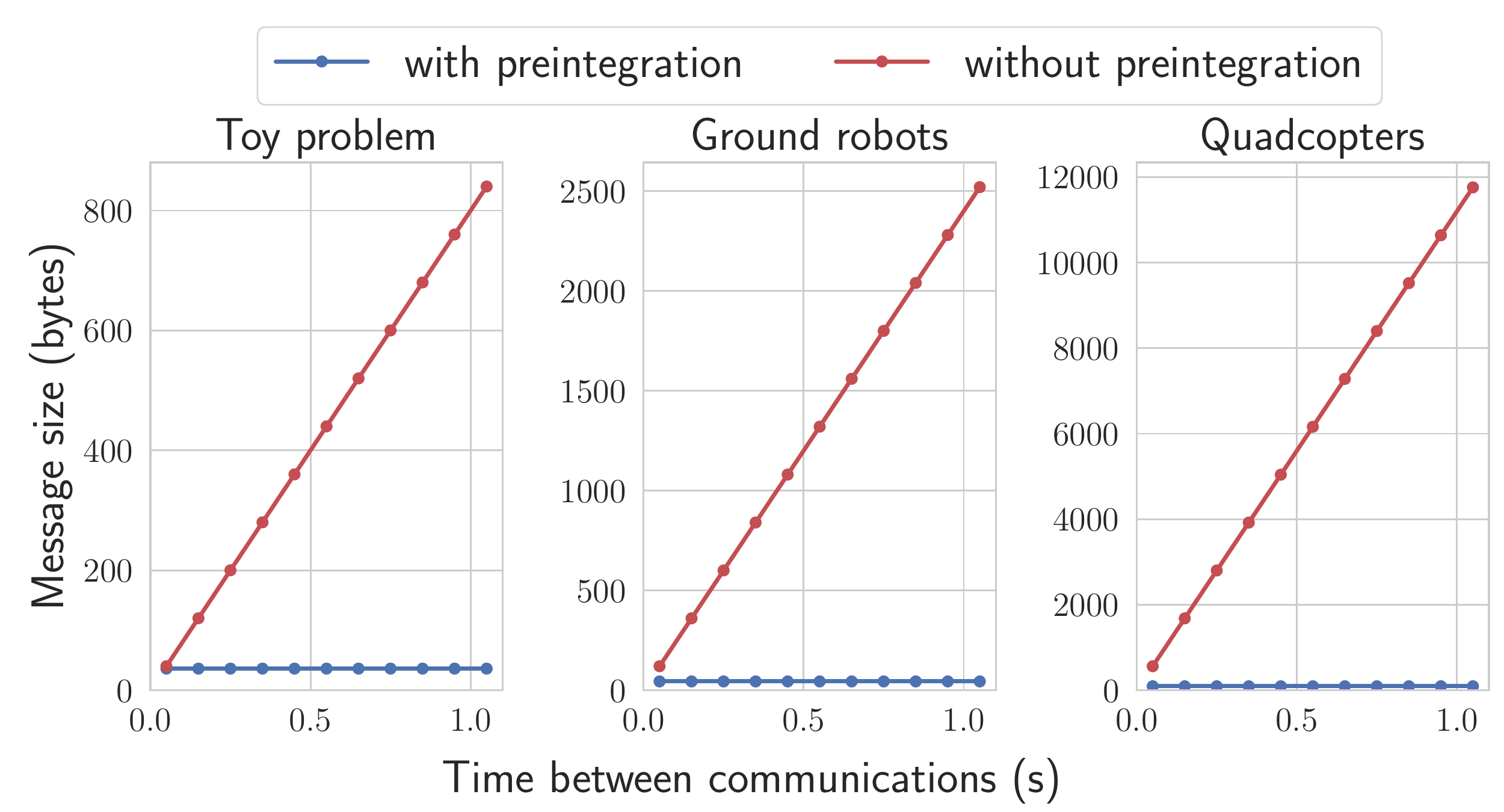}
    \caption{\diff{Message size in bytes required to share odometry, as a function of the time period between communications between two robots. The red line ``without preintegration'' naively transmits all input measurements that occured within the time period. Preintegration maintains a constant message size while providing identical information.}}
    \label{fig:rmi_packet_size}
\end{figure}

Concretely, using IMU preintegration as an example, the RMI itself must be communicated, which requires 10 floating-point numbers. However, the covariance matrix is $15 \times 15$, which would require communication of an additional 120 floating-point numbers to represent one of its triangular halves. With the proposed autoencoder, these 120 numbers are replaced with an encoding consisting of \emph{one} number, thus dramatically reducing the communication cost. \diff{Figure \ref{fig:rmi_packet_size} shows how the required communication rate varies with the duration between two successive communications between two arbitary robots. The naive solution without preintegration requires sharing all input measurements that have occured during that period, whereas preintegration yields a constant message size. The proposed method can be applied to all problems discussed in the paper, where the networks must be trained for each problem.}

\section{Simulation with Ground Robots} \label{sec:ground_robots}

The proposed algorithm is tested in a simulation with ground robots, shown in the center of Figure \ref{fig:examples}. Each robot estimates their own pose and their neighbors' poses relative to a world frame. Denoting the pose of Robot $i$ relative to the world frame $w$ as $\mbf{T}_{wi} \in SE(2)$, the state of an arbitrary robot is given by 
\beq 
\mc{X}_{i_k} = \left(\mbf{T}^{[i]}_{wi_k}, \mbf{T}^{[i]}_{wj_k}, \ldots\right),\qquad  j \in \mc{N}_i,
\eeq 
where, again, the $(\cdot)^{[i]}$ superscript indicates Robot $i$'s estimate or ``instance'' of that physical quantity. Each robot collects wheel odometry at 100~Hz, providing $\mbf{u}_{i_k} = [\omega_{i_k}\; v_{i_k}\; 0]^\trans$ as input measurements, where $\omega_{i_k}$ is Robot $i$'s angular velocity and $v_{i_k}$ is its forward velocity in its own body frame. The pose kinematics for any single robot along with its preintegration are shown in Example \ref{ex:wheel_odometry}. When Robot $i$ receives an input measurement, it updates the part of its state corresponding to its own pose to create 
\beq 
\mathfrak{X}_{i_k} = \left(\mbf{T}^{[i]}_{iw_{k-1}} \Exp(\Delta t \mbf{u}_{i_{k-1}}), \mbf{T}^{[i]}_{wj_{k-1}}, \ldots \right).
\eeq 
The neighbor poses $\mbf{T}_{wj_{k-1}}$ are now out of date, as neighboring odometry information is not yet accessible to Robot $i$, and this partially out-of-date state is non-physical and given the symbol $\mathfrak{X}_{i_k}$. \diff{Every robot computes their own RMIs from wheel odometry using the equations from Example \ref{ex:wheel_odometry}.} When a neighbor's RMI $ \Delta \mbf{T}_{j_{pq}}$ is received at some later time step $k=q$, the state is updated with 
\beq 
\mc{X}_{i_{q}} = \left(\mbf{T}^{[i]}_{iw_q}, \mbf{T}^{[i]}_{wj_p} \Delta \mbf{T}_{j_{pq}}, \ldots \right)
\eeq
where $p$ represents the time step index of the last time a neighbor RMI was received. 

Each robot also collects range measurements to its neighbors at 10~Hz, \diff{with the connectivity graph shown in Figure \ref{fig:examples} (middle).} Only two robots collect relative position measurements to known landmarks at 10~Hz. At an arbitrary separate frequency, each robot sends its current state and covariance to its neighbors, allowing the neighbors to compute pseudomeasurements of the form 
\beq 
\mbf{c}_{ij}(\mc{X}_{i}, \mc{X}_{j}) = \bma{c} \Log\left(\mbf{T}_{wi}^{[i]^{-1}} \mbf{T}_{wi}^{[j]}\right) \\ \Log\left(\mbf{T}_{wj}^{[i]^{-1}}  \mbf{T}_{wj}^{[j]}\right) \\  \Log\left(\mbf{T}_{w\ell}^{[i]^{-1}} \mbf{T}_{w\ell}^{[j]}\right) \\ \vdots\ema, \qquad \ell \in \mc{N}_i \cap \mc{N}_j.
\eeq
A simulation is performed with 4 robots each executing Algorithm \ref{alg:alg2}, with root-mean-squared error (RMSE) shown in Figure \ref{fig:se2_rmse}. \diff{The initial states are initialized to ground truth with some random error with covariance $\mbfcheck{P}_{i_0} =0.1^2 \cdot \mbf{1}$. The position and range measurements have Gaussian noise with $0.3$~m and $0.1$~m of standard deviation, respectively, and the pseudomeasurement covariance is $\mbs{\Psi}=\mbf{0}$.} The results show that all four robots' estimation errors successfully stabilize and remain low, despite only two robots having sufficient sensors to make their states observable. \diff{Similar to the Toy Problem, the naive solution is implemented which is identical to the proposed solution, but does not do a CI step.} \diff{A 50-trial Monte Carlo simulation was also performed with the resulting average NEES plotted in Figure \ref{fig:se2_nees}. The naive solution has significantly higher error than both the centralized or proposed algorithms and is so overconfident that it cannot be plotted within reasonable axis limits in Figure \ref{fig:se2_nees}. Although, in theory, the proposed algorithm should have lower NEES values than the centralized solution, Figure \ref{fig:se2_nees} shows comparable values. This is suspected to be due to the high degree of nonlinearity of the problem.}

\diff{As seen in Figure \ref{fig:se2_rmse}}, if state fusion is done at a sufficiently high frequency, performance is even comparable to the centralized estimator, \diff{but this will incur a larger communication and computation requirement as discussed in Section \ref{sec:general_problem}.}

\begin{figure*}[t]
    \centering
    \includegraphics[width=\linewidth]{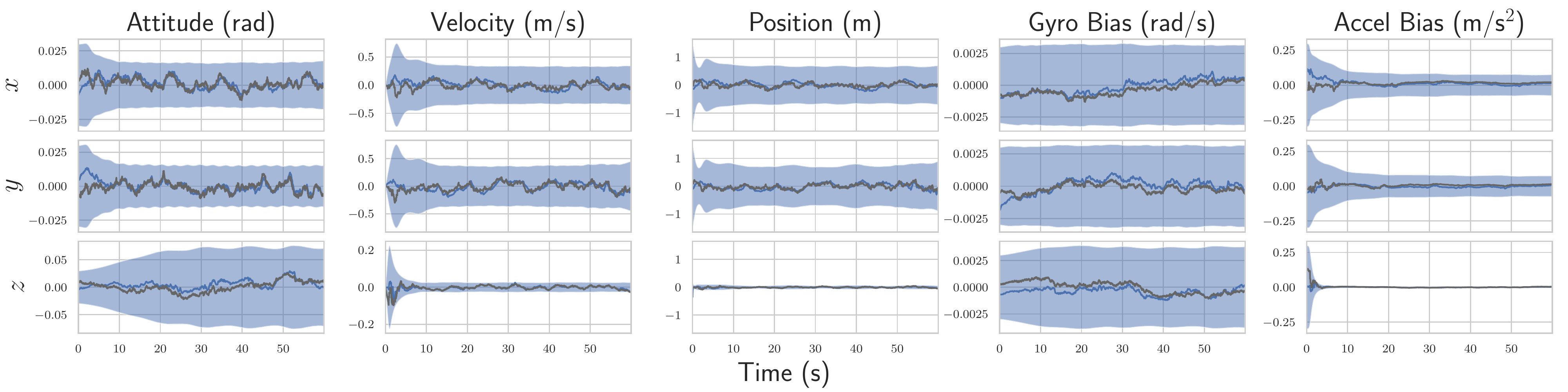}
    \caption{Simulated estimation error of Robot 3's estimate of its own kinematic state and IMU biases. The estimate and corresponding bounds with the proposed algorithm are shown in blue, with the centralized estimate overlayed in dark grey. For attitude, the $x-y-z$ components represent roll-pitch-yaw errors respectively. Note that Robot 3 does not have position measurements, and therefore cannot observe the states shown in this plot without information sharing. The naive solution has rapidly diverging error and is not plotted.}
    \label{fig:quad_sim}
\end{figure*}

\section{Simulation and Experiments with Quadcopters} \label{sec:quads}

To demonstrate the flexibility of the proposed framework, consider a new problem involving quadcopters. The kinematic state of each quadcopter is modelled using extended pose matrices $\mbf{T} \in SE_2(3)$ (\cite{Brossard2020b}). Each robot estimates both their absolute pose relative to the world frame $\mbf{T}_{wi} \in SE_2(3)$, their own IMU bias $\mbf{b}_i$, as well as the \emph{relative} poses of their neighbors $\mbf{T}_{ij} \in SE_2(3)$ and their IMU bias $\mbf{b}_j$. The full state of Robot $i$ is then given by 
\beq 
\mc{X}_i = (\mbf{T}_{wi}, \mbf{b}_i^{[i]}, \mbf{T}_{ij}, \mbf{b}_j^{[i]}, \ldots),\qquad j \in \mc{N}_i.
\eeq 
The pose of Robot $j$ relative to Robot $i$ $\mbf{T}_{ij}$ has kinematics involving the IMU measurements of both robots, and are given in discrete time by 
\beq 
\mbf{T}_{ij_k} = \mbf{U}_{i_{k-1}}^{-1} \mbf{T}_{ij_{k-1}} \mbf{U}_{j_{k-1}},
\eeq
where $\mbf{U}_{j_{k-1}}$ has an identical definition as in \eqref{eq:imu_process}, but computed from Robot $j$'s IMU measurements. When Robot $i$ receives input measurements from its own IMU $\mbf{u}_k$, it predicts the part of its own state corresponding to its own pose, and additionally performs a partial prediction on the relative poses with 
\beq 
\mathfrak{X}_{i_k} = (\mbf{G}_{k-1}\mbf{T}_{wi_{k-1}}\mbf{U}_{i_{k-1}}, \mbf{b}_{i_{k-1}}^{[i]}, \mbf{U}_{i_{k-1}}^{-1}\mbf{T}_{ij_{k-1}}, \mbf{b}_{j_{k-1}}^{[i]}, \ldots).
\eeq 
The terms $\mathfrak{T}_{ij_{k-1}} \triangleq \mbf{U}_{i_{k-1}}^{-1}\mbf{T}_{ij_{k-1}}$, which are the partially-predicted neighbor poses, are  a strange, non-physical intermediate state. Only when the neighbor's RMI $\Delta \mbf{U}_{j_{pq}}$ is received do the neighbor poses regain meaning with $ \mbf{T}_{ij_q} = \mathfrak{T}_{ij_{p}} \Delta \mbf{U}_{j_{pq}}$. However, since biases are also being estimated in this problem, Robot $i$ must first correct the neighbor's raw RMIs $\Delta \mbf{U}_{j_{pq}}(\mbf{u}_{j_{p:q-1}},\mbf{0})$ using its estimate of the neighbor's IMU bias, as described in Section \ref{sec:biases}. That is,
\beq 
\Delta \mbf{U}_{j_{pq}} \approx \Delta \mbf{U}_{j_{pq}}(\mbf{u}_{j_{p:q-1}}, \mbf{0}) \oplus \mbf{B}_{j_{pq}} \mbf{b}_{j_q}^{[i]},
\eeq
leading to the full state update given by 
\beq 
\mc{X}_{i_q} = (\mbf{T}_{wi_{q}}, \mbf{b}^{[i]}_q,  \mathfrak{T}_{ij_{p}} \Delta \mbf{U}_{j_{pq}}, \mbf{b}_{j_q}^{[i]}\ldots).
\eeq

Finally, the pseudomeasurements chosen for this problem are 
\beq 
\mbf{c}_{ij}(\mc{X}_{i}, \mc{X}_{j}) = \bma{c} \Log\left(\mbf{T}_{wi} \mbf{T}_{ij} \mbf{T}_{wj}^{-1} \right) \\ \Log\left(\mbf{T}_{ij} \mbf{T}_{ji}\right) \\ \mbf{b}^{[i]}_i - \mbf{b}^{[j]}_i \\ \mbf{b}^{[i]}_j - \mbf{b}^{[j]}_j \\ \Log\left( \mbf{T}_{ij}\mbf{T}_{j \ell} \mbf{T}_{i\ell} ^{-1}\right) \\ \vdots \ema , \qquad \ell \in \mc{N}_i \cap \mc{N}_j.
\eeq
with corresponding covariance $\mbs{\Psi} = \mbf{0}$.
\begin{figure}
    \centering 
    \includegraphics[width=0.49\linewidth]{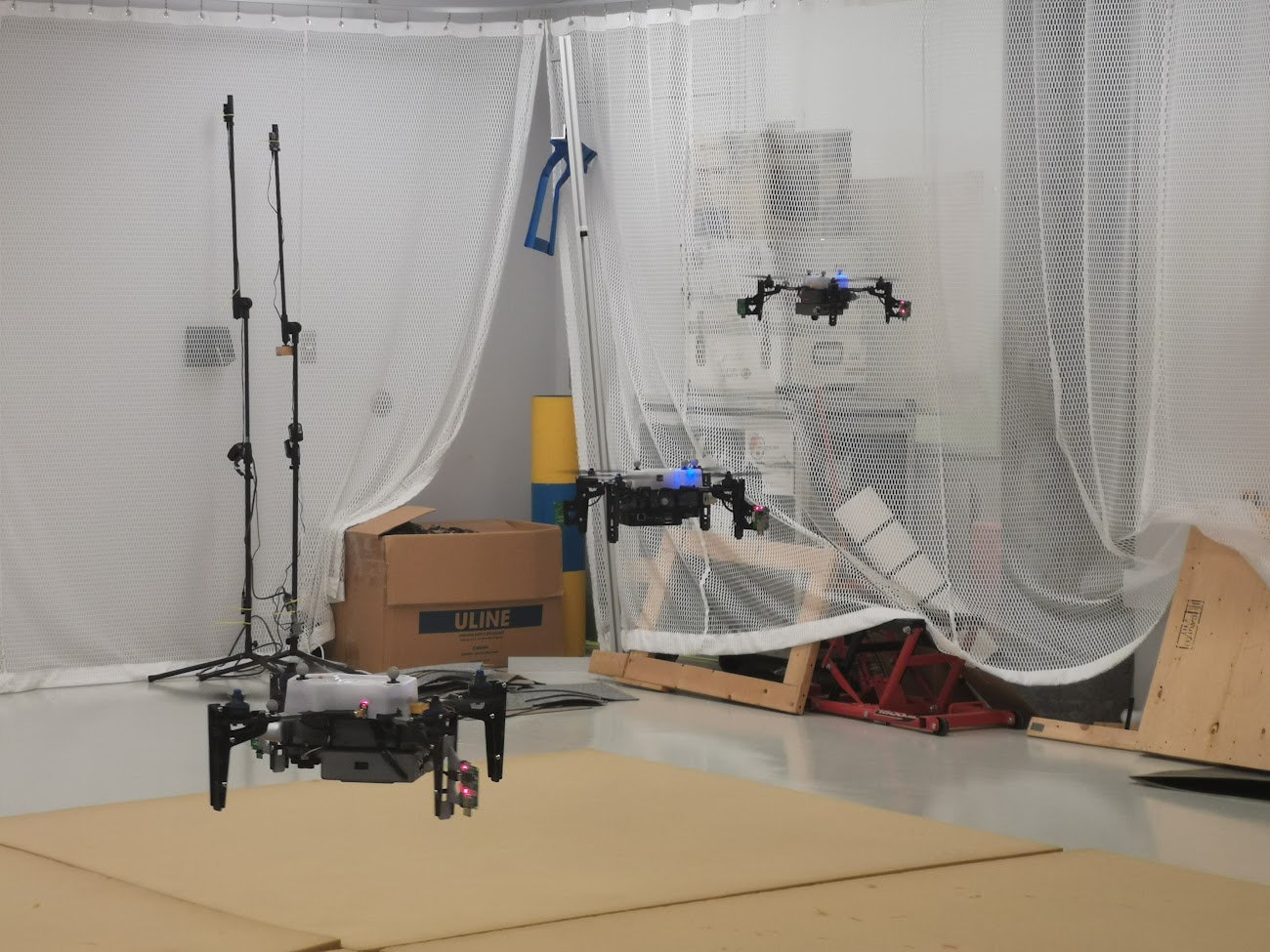}\hspace{0.1cm}%
    \includegraphics[width=0.49\linewidth,trim={4.87cm 3cm 5cm 0cm}, clip]{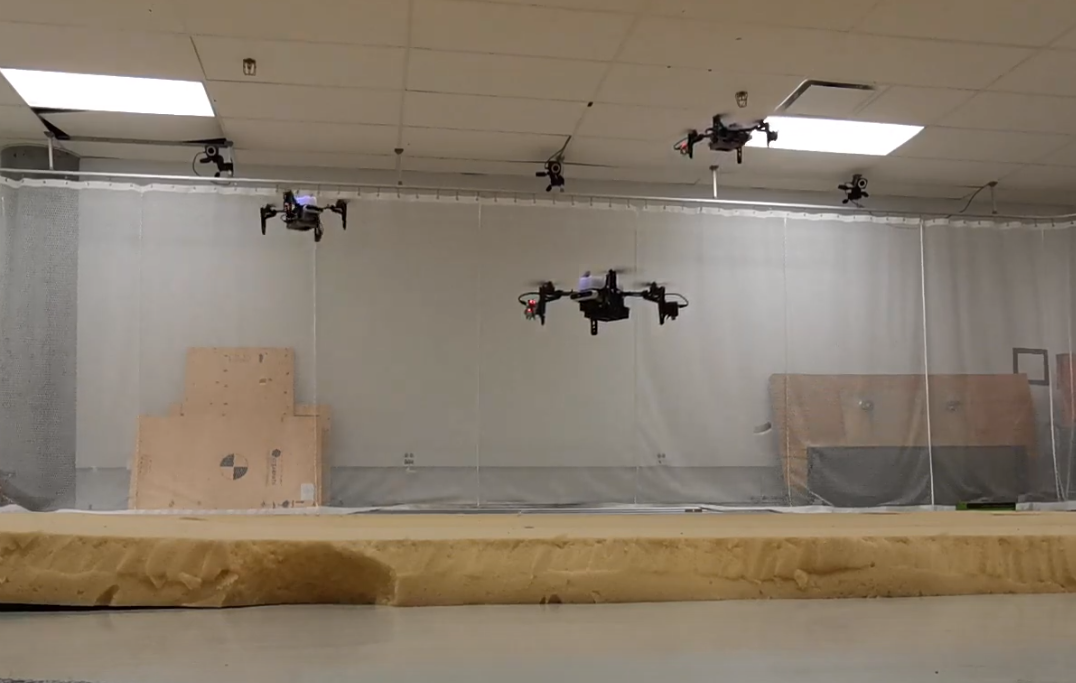} \\
    \vspace{0.1cm}
    \includegraphics[width=0.49\linewidth]{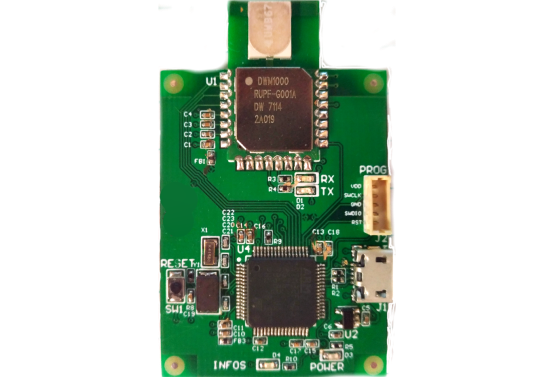}\hspace{0.1cm}%
    \includegraphics[width=0.49\linewidth]{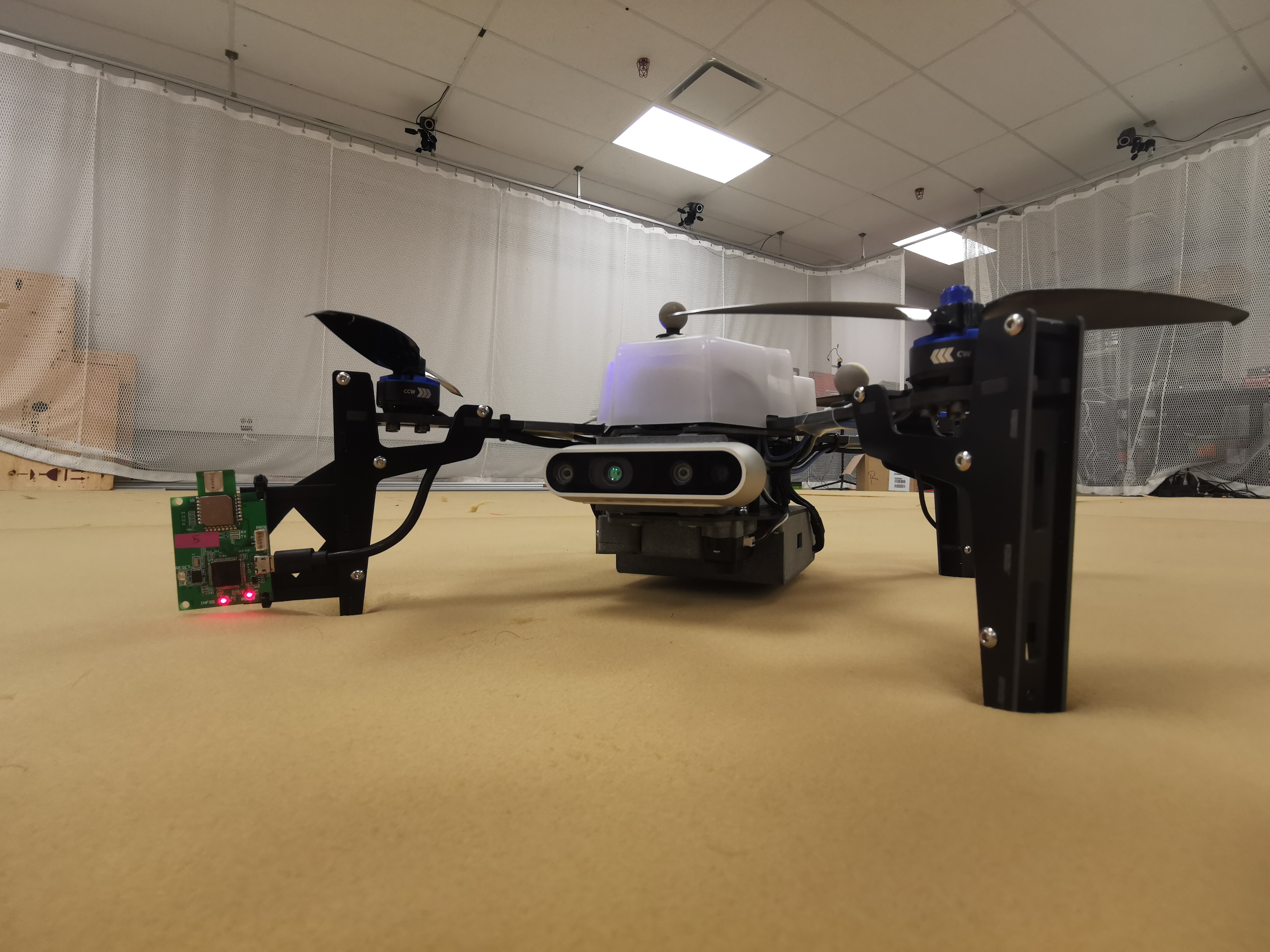}
    \caption{\textbf{Top:} Three quadcopters in flight under a motion capture system. \textbf{Bottom left:} custom UWB module. \textbf{Bottom right:} a close up of the Uvify IFO-S quadcopter, fitted with a UWB module seen on the left leg, as well as on the opposite leg. }
    \label{fig:hardware}
\end{figure}
\begin{figure}
    \centering 
    \includegraphics[width=0.5\linewidth,clip,trim={3.5cm 1.7cm 1.5cm 2.9cm}]{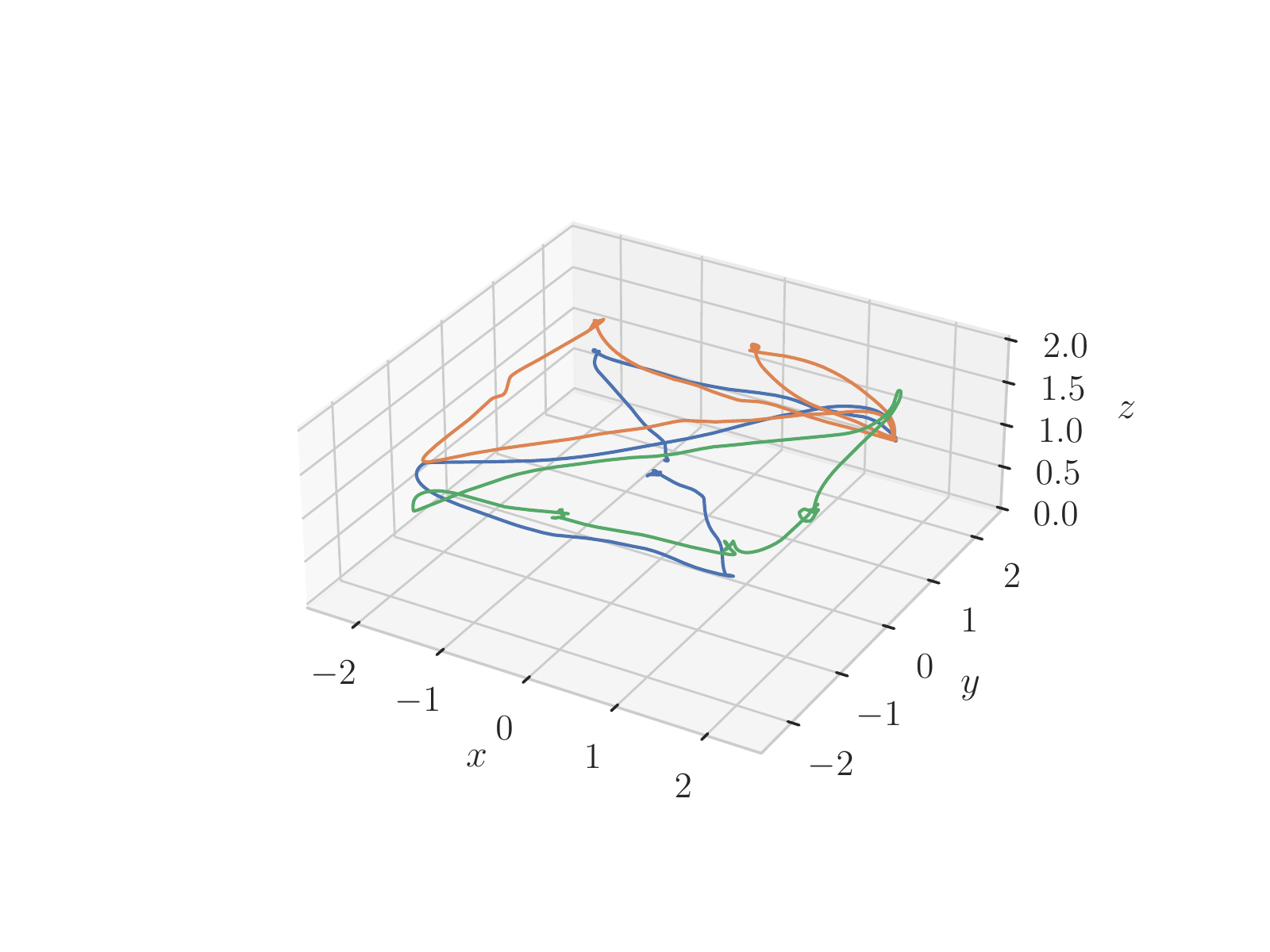}%
    \includegraphics[width=0.5\linewidth,clip,trim={3.5cm 1.7cm 1.5cm 2.9cm}]{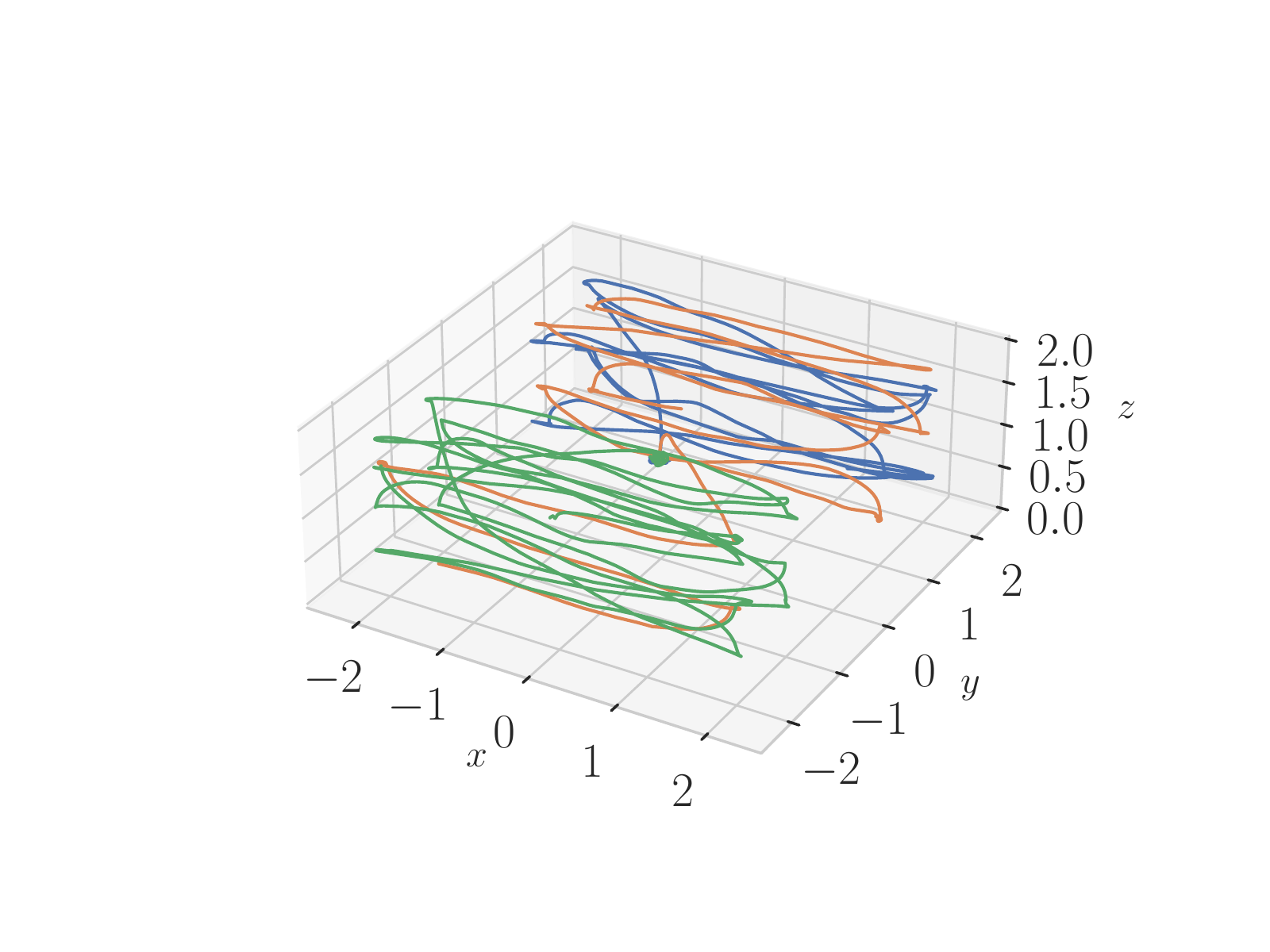}
    \caption{Examples of the various trajectories flown in the experimental trials, where each color represents a different quadcopter.}
    \label{fig:trajectories}
\end{figure}
\begin{table*}[t]
    \centering
    \caption{Self- Positioning RMSE (m) from experimental trials.}
    \label{tab:experiments}
    \begin{tabular}{@{}cccccccccc@{}}
        \toprule
        \multicolumn{1}{c}{\multirow{2}{*}{Trial \#}} & \multicolumn{3}{c}{Centralized} & \multicolumn{3}{c}{Proposed} & \multicolumn{3}{c}{Error Reduction}  \\ \cmidrule(lr){2-4}\cmidrule(lr){5-7}\cmidrule(lr){8-10}
        \multicolumn{1}{c}{} & Robot 1 & Robot 2 & Robot 3 & Robot 1 & Robot 2 & Robot 3 & Robot 1 & Robot 2 & Robot 3  \\ \midrule
        1 &  0.43 & 0.49 & 0.55 & 0.22 & 0.22 & 0.61 &  -48\%& -54\% & 10\%  \\ 
        2 &  0.18& 0.26 & 0.34 & 0.16 & 0.18 & 0.40 & -13\% & -28\% & 16\%  \\ 
        3 &  0.17& 0.24 & 0.68 & 0.16 & 0.17 & 0.45 & -10\% & -28\% &  -33\%  \\ 
        4 &  0.20& 0.25 & 0.31 & 0.26 & 0.28 & 0.48 & 32\% & -13\% &   55\%\\ \bottomrule
        \end{tabular}
\end{table*}

\subsection{Hardware Setup}
The hardware setup in these experiments can be seen in Figure \ref{fig:hardware}. Three Uvify IFO-S quadcopters are used that each possess an IMU at 200~Hz, a 1D LIDAR height sensor at 30~Hz, and magnetometers at 30~Hz. Additionally, two ultra-wideband (UWB) transceivers are installed on the quadcopter legs, producing inter-robot distance measurements at 90~Hz for each robot. \diff{As shown by \cite{Shalaby2021}, installing multiple UWB tags per robot 
\ms{results in}
relative position observability.} The UWB transceivers are custom-printed modules that use the DW1000 UWB transceiver. The firmware for these modules has been written in C, implementing a double-sided two-way-ranging protocol with details described by \cite{Shalaby2022a}. \cite{Shalaby2022a} also describe the power-based bias calibration and noise characterization procedure used in these experiments. Since all transceivers operate on the same frequency in these experiments, only one can transmit at a time to avoid interference. A decentralized scheduler is therefore implemented that continuously cycles through all transceiver pairs one at a time, obtaining range measurements and potentially transmitting other useful data. \diff{In these experiments, the communication graph is complete with all quadcopters capable of communicating with each other. Preintegrated RMIs are shared whenever a UWB measurement occurs, and state sharing occurs at a separate frequency of 10Hz.}

A Vicon motion capture system is used to collect ground truth, from which synthesized absolute position measurements with a standard deviation of 0.3 m are generated for Robots 1 and 2 only. Robot 3 does not receive absolutely position measurements, nor any magnetometer measurements, and therefore has no absolute pose information available without communication with the other two robots. Example trajectories for some of the experimental trials are shown in Figure \ref{fig:trajectories}.

\subsection{Simulation Results}
The algorithm is first tested with simulated versions of the described quadcopters, and the estimation results for Robot 3's absolute pose and bias are shown in Figure \ref{fig:quad_sim}. Although there are many other states associated with the simulation, these states are the most interesting as they are the ones that are unobservable without incorporation of the pseudomeasurements. Figure \ref{fig:quad_sim} shows that Robot 3 is capable of estimating its own absolute pose and bias, using information from sensors located on Robots 1 and 2. Furthermore, the errors remain within the 3-sigma confidence bounds, even with the first-order RMI bias correction, indicating statistical consistency.

\begin{figure}
    \centering
    \includegraphics[width=\linewidth]{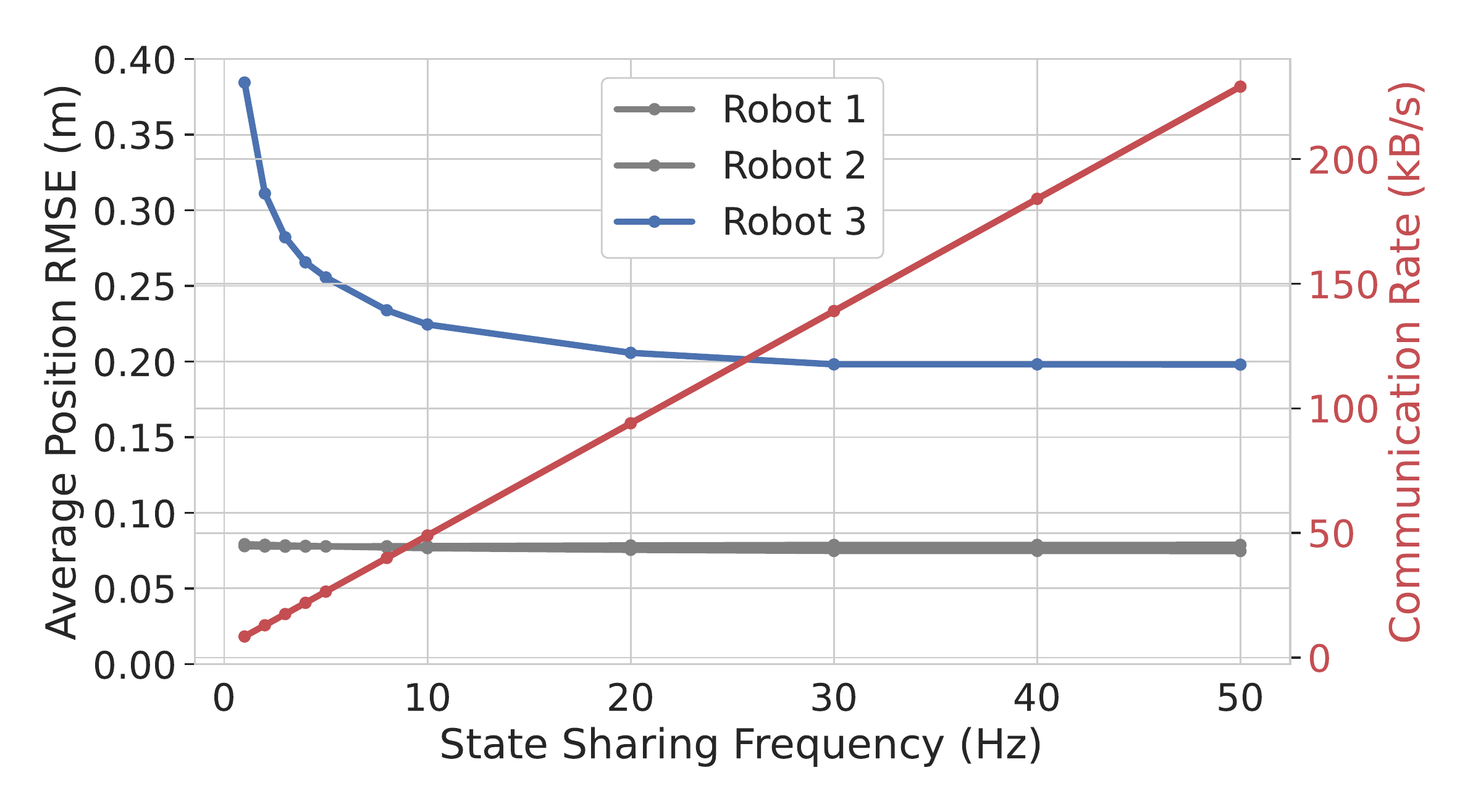}
    \caption{\diff{Average self-positioning RMSE with varying communication rate for the simulated version of the quadcopter problem. Robot 3 does not receive position measurements, and hence is reliant on the other robots to have an observable state.}}
    \label{fig:ablation}
\end{figure}

\diff{Figure \ref{fig:ablation} shows the positioning RMSE for varying frequency at which state information between robots is shared. At lower frequencies, Robot 3's estimate has more time to drift between communications, and hence there is higher error. For this problem, roughly the same estimation performance is achieved for state sharing of 20~Hz and above, with 10~Hz being a compromising value providing a trade-off between accuracy and communication cost.}

\subsection{Experimental Results}
\begin{figure}
    \centering 
    \includegraphics[width=\linewidth,clip,trim={0.2cm 0.7cm 0.2cm 0cm}]{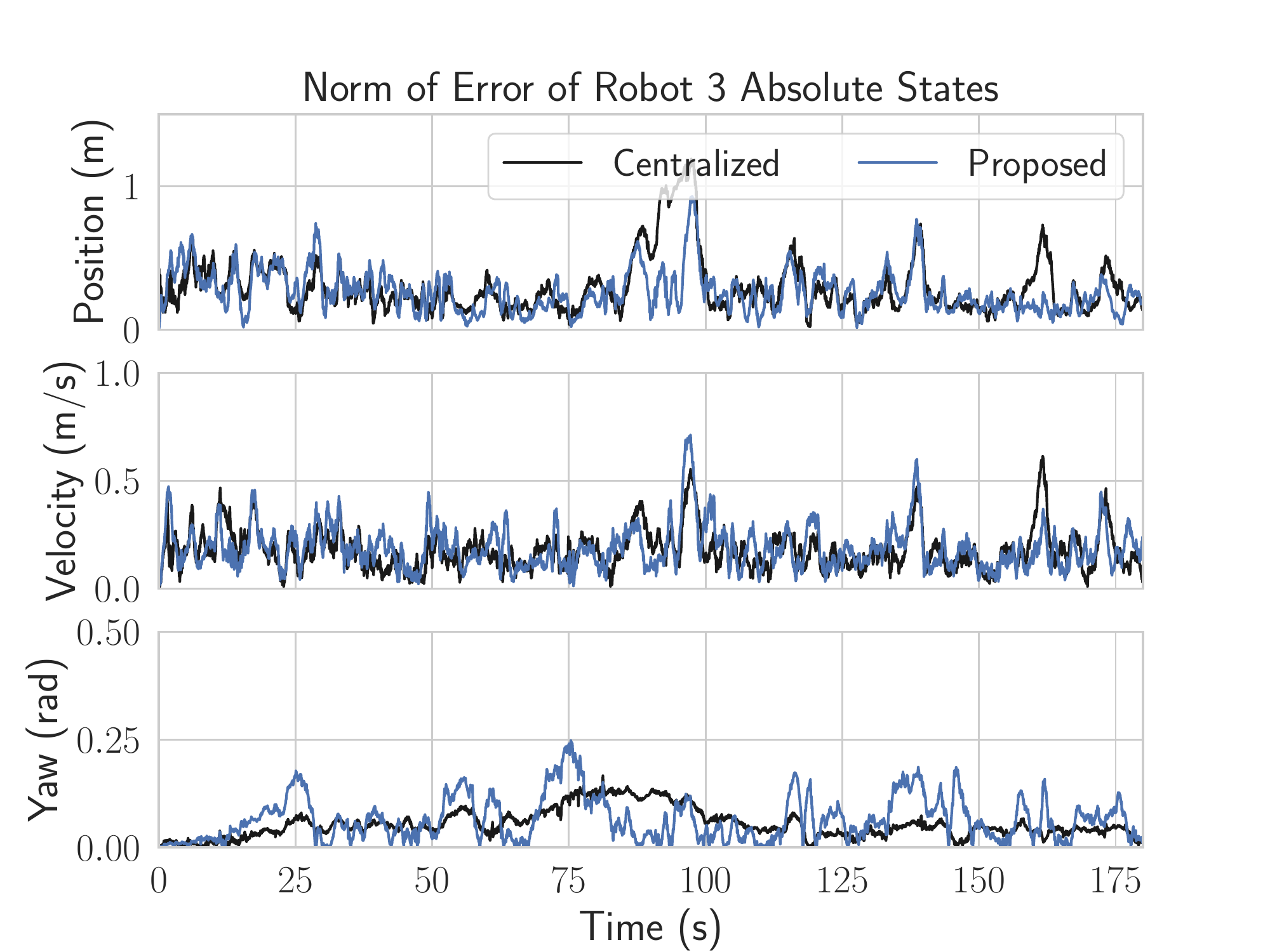}%
    \caption{Position, velocity, and yaw RMSE for Robot 3 from one of the experimental trials. Since Robot 3 has no position measurements, these quantities are unobservable without the fusion of pseudomeasurements.}
    \label{fig:experiment_rmse}
\end{figure}

\begin{figure}
    \centering 
    \includegraphics[width=\linewidth,clip,trim={0.2cm 0.7cm 0.2cm 0cm}]{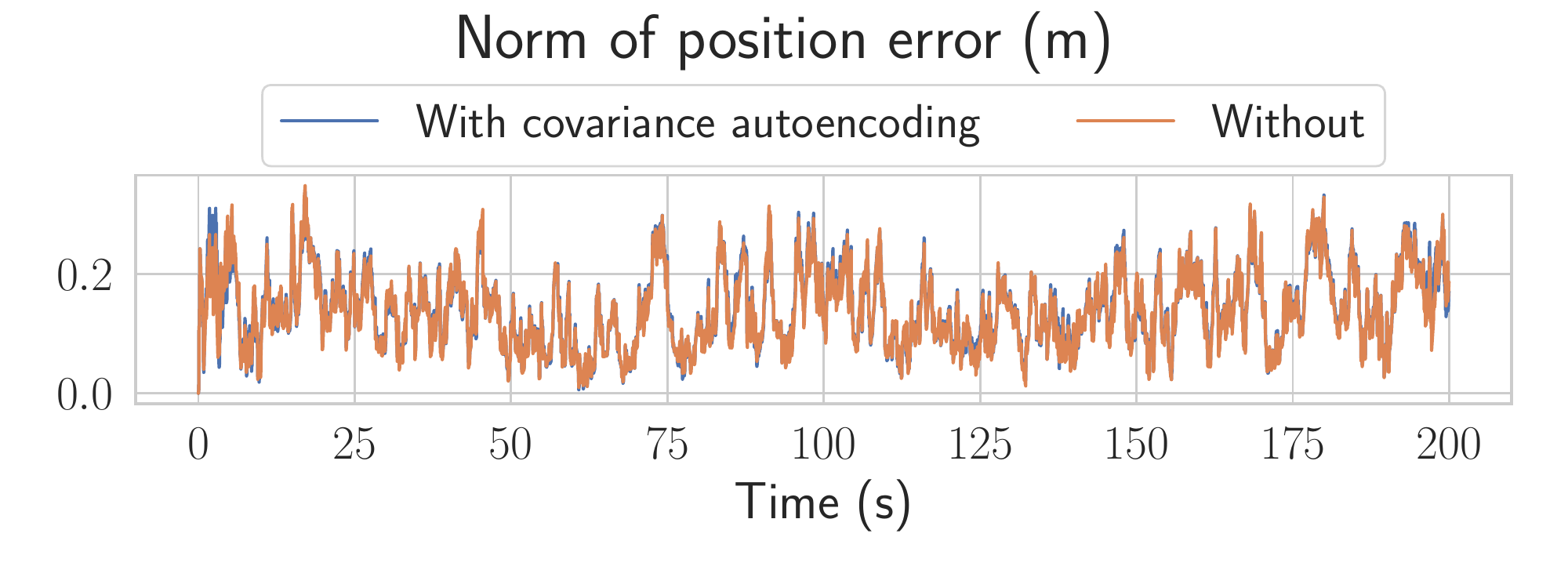}%
    \caption{The effect of preintegrated covariance autoencoding, as described in Section \ref{sec:preint_autoencoding}, on the position estimate of Robot 1. The two lines are almost identical, showing that the proposed autoencoder induces minimal error on the estimate. All other states have similar plots.}
    \label{fig:autoencoder_estimator}
\end{figure}
Multiple experimental runs are performed on different days, with the absolute positioning results for each robot viewable in Table \ref{tab:experiments}. In some cases, the proposed algorithm even outperforms the centralized solution, which is theoretically optimal. However, the real world contains many unmodelled sources of error, such as frame misalignments, timestamping errors, vibrations, and UWB ranging outliers. \diff{These effects may break the assumptions that the optimality of the centralized estimator relies on. Even after tuning covariances to obtain the best centralized performance}, it appears that the results benefit from the covariance inflation resulting from CI. For both estimators, the IMU is calibrated to compensate for large biases and scaling factors. The normalized-innovation-squared test (\cite{Bar-Shalom2001}) is also used to reject UWB outliers in both estimators,

A plot of RMSE versus time for Robot 3's absolute states can be seen in Figure \ref{fig:experiment_rmse}, which are states that are unobservable from Robot 3's own measurements. Again, Figure \ref{fig:experiment_rmse} shows that error magnitudes lie in similar ranges for both the centralized and decentralized estimators. Figure \ref{fig:autoencoder_estimator} compares two decentralized estimator runs, with one using the mean-assisted autoencoder from Section \ref{sec:preint_autoencoding}. As desired, the lines are identical, and the plot shows that the estimate is unaffected by the autoencoding. This means that the autoencoder is highly effective at compressing the covariance matrix with minimal reconstruction error.

\section{Conclusion}
This paper presents a general-purpose algorithm for decentralized state estimation in robotics. The algorithm is the result of a new way to formulate the decentralized state estimation problem, specifically with the assistance of pseudomeasurements that allow the definition of arbitrary nonlinear relationships between robot states. \diff{For problems involving relative measurements, a communication-efficient approach is proposed for preintegratable process models, as defined by \eqref{eq:preint3}, where state-change information is shared in the form of relative motion increments.}  The algorithm is tested on three different problems, each involving a variety of state definitions, process models, and measurements. \diff{In all of the presented problems, robots only need to share their states, RMIs, and corresponding covariances, which ultimately results in average transmission rates per robot of 0.2~kB/s for the toy problem, 4.5~kB/s for the ground robots, and 53.2~kB/s for the quadcopters.}

Thanks to covariance intersection, the algorithm is appropriate for arbitrary graphs, and does not require any bookkeeping, growing memory, buffering of measurements, or reprocessing of data. At the same time, the approximation made by covariance intersection makes the proposed method suboptimal, as it is well-known to be overly-conservative. Nevertheless, in the specific problems shown in this paper, the results using CI have been satisfactory provided that the fusion frequency is high enough, and the communication graph is not too sparse. It is also worth mentioning that the proposed algorithm still assumes that process model inputs, whether in raw or preintegrated form, have noise that is uncorrelated with the robot states, just like the sensor measurement noise. These assumptions must hold for a consistent estimator. In this paper, it is only correlations between different robots' states that are mitigated by covariance intersection.

\diff{One limitation of this proposed approach is that the communication cost grows quadratically with the state size, since the state covariance is also shared. While this is not an issue for small state sizes, such as those representing 3D poses, it could become a problem for states involving multiple time steps or a very large number of robots. Furthermore since this paper allows for variable state definitions between robots, the state definition itself for each robot may need to be communicated, or established a priori. The initial value of the state is also assumed to be known through an arbitrary initialization procedure.}

Future work can consider improving the approximation made by CI and compressing the covariance matrix associated with the state. Also, using the proposed MAP approach with pseudomeasurements, it should be possible to derive decentralized batch and sliding-window estimators, often termed \emph{smoothers}, since these algorithms also originate from the MAP problem.

\begin{funding}
  This work was supported by the NSERC Alliance Grant program, the NSERC Discovery Grant Program, the CFI JELF program, and by the FRQNT.
\end{funding}

\begin{acks}
  The authors would like to thank Justin Cano and J\'er\^ome Le Ny for their assistance in the development of custom UWB modules.
\end{acks}
  
\bibliographystyle{SageH}
\bibliography{library.bib}

\end{document}